\documentclass[conference]{IEEEtran}
\usepackage{cite}
\usepackage[usenames, dvipsnames]{color}
\usepackage{xspace}
\usepackage{booktabs}
\usepackage{enumerate}
\usepackage{verbatim}
\usepackage{amsmath}
\usepackage{amsfonts}
\usepackage{graphicx}
\usepackage{subcaption}
\usepackage[linesnumbered,ruled]{algorithm2e}
\usepackage{siunitx}
\usepackage{float}
\usepackage{multirow}
\usepackage[colorlinks=true,urlcolor=black,linkcolor=red,citecolor=red]{hyperref}

\newif\ifdraft
\draftfalse

\newif\ifpublishable
\publishabletrue

\ifdraft
  \newcommand{\todocolor}[1]{\textcolor{red}{#1}}
\else
  \newcommand{\todocolor}[1]{}
\fi
\newcommand{\mahmood}[1]{\todocolor{[[Mahmood: #1]]}}
\newcommand{\lujo}[1]{\todocolor{[[Lujo: #1]]}}
\newcommand{\mike}[1]{\todocolor{[[Mike: #1]]}}

\newcommand{\secref}[1]{\mbox{Sec.~\ref{#1}}\xspace}

\newcommand{\figref}[1]{\mbox{Fig.~\ref{#1}}}
\newcommand{\tabref}[1]{\mbox{Table~\ref{#1}}}

\newcommand{\eqnref}[1]{Eqn.~\ref{#1}\xspace}

\makeatletter
\DeclareRobustCommand{\varname}[1]{\begingroup\newmcodes@\mathit{#1}\endgroup}
\makeatother

\newcommand{\nML}[1][\nvariants]{#1-ML}
\newcommand{\ml}[0]{ML}
\newcommand{\dnn}[0]{DNN}
\newcommand{\lpnorm}[1]{$L_{#1}$}
\newcommand{\pgd}[0]{$\mathit{PGD}$}
\newcommand{\advpgd}[0]{$\mathit{Adv}_{\mathit{PGD}}$}
\newcommand{\lid}[0]{$\mathit{LID}$}
\newcommand{\nic}[0]{$\mathit{NIC}$}
\newcommand{\dnnLogits}[0]{\mathbb{L}}
\newcommand{\cwloss}[0]{\textit{Loss}_\textit{cw}}
\newcommand{\cwlosstarg}[0]{\textit{Loss}_\textit{cw}^\textit{targ}}
\newcommand{\cwlossuntarg}[0]{\textit{Loss}_\textit{cw}^\textit{untarg}}
\newcommand{\ocsvm}[0]{\textit{oc-SVM}}
\newcommand{\celoss}[0]{\textit{Loss}_\textit{ce}}
\newcommand{\nclasses}[0]{m}

\newcommand{\nvariants}[0]{\ensuremath{n}\xspace}
\newcommand{\threshold}[0]{\tau}
\newcommand{\nMLFunc}[0]{\mathbb{F}}
\newcommand{\dnnFunc}[1]{\mathbb{F}_{#1}}
\newcommand{\mnist}[0]{MNIST}
\newcommand{\gtsrb}[0]{GTSRB}
\newcommand{\cifarten}[0]{CIFAR10}
\newcommand{\wresnet}[2]{Wide-ResNet-#1-#2}
\newcommand{\msr}[0]{\emph{MSR}}
\newcommand{\tsr}[0]{\emph{TSR}}

\newcommand{\parheading}[1]{\textbf{\textit{#1}}~\hspace{2pt}}

\begin{document}

\title{\nML{}: Mitigating Adversarial Examples via Ensembles of Topologically
Manipulated Classifiers}

\ifpublishable
\IEEEoverridecommandlockouts
\author{\IEEEauthorblockN{Mahmood Sharif\textsuperscript{\dag}}\thanks{\textsuperscript{\dag}Work done as a Ph.D.\ student at Carnegie Mellon University.}
\IEEEauthorblockA{NortonLifeLock Research Group\\
\texttt{\url{mahmood\_sharif@symantec.com}}}
\and
\IEEEauthorblockN{Lujo Bauer}
\IEEEauthorblockA{Carnegie Mellon University\\
\texttt{\url{lbauer@cmu.com}}}
\and
\IEEEauthorblockN{Michael K. Reiter}
\IEEEauthorblockA{University of North Carolina at Chapel Hill\\
\texttt{\url{reiter@cs.unc.edu}}}}
\fi

\maketitle

\begin{abstract}
  This paper proposes a new defense called \nML{} against adversarial
  examples, i.e., inputs crafted by perturbing benign inputs by small
  amounts to induce misclassifications by classifiers.  Inspired by
  \nvariants-version programming, \nML{} trains an ensemble of
  \nvariants classifiers, and inputs are classified by a vote of the
  classifiers in the ensemble.  Unlike prior such approaches, however,
  the classifiers in the ensemble are trained specifically to classify
  adversarial examples differently, rendering it very difficult for an
  adversarial example to obtain enough votes to be misclassified.  We
  show that \nML{} roughly retains the benign classification accuracies of
  state-of-the-art models on the \mnist{}, \cifarten{}, and \gtsrb{}
  datasets, while simultaneously defending against adversarial
  examples with better resilience than the best defenses known to date
  and, in most cases, with lower classification-time overhead.
\end{abstract}

\section{Introduction}

Adversarial examples---minimally and adversarially perturbed variants of benign
samples---have emerged as a challenge for machine-learning (\ml{}) algorithms
and systems. Numerous attacks that produce inputs that evade correct classification by \ml{} algorithms, and
particularly deep neural networks (\dnn{}s), at inference time have been
proposed (e.g.,~\cite{Biggio13Evasion, Gdfllw14ExpAdv, Szegedy13NNsProps}).
The attacks vary in the perturbation types that they 
allow and the application domains, but they most commonly focus on
adversarial perturbations with bounded \lpnorm{p}-norms to evade \ml{} algorithms for
image classification (e.g.,~\cite{Biggio13Evasion, Carlini17Robustness,
  Gdfllw14ExpAdv, Papernot16Limitations, Szegedy13NNsProps}). Some attacks 
minimally change physical artifacts to mislead recognition \dnn{}s (e.g., adding
patterns on street signs to mislead street-sign
recognition)~\cite{Evtimov17Signs, Sharif16AdvML}. Yet others imperceptibly perturb audio
signals to evade speech-recognition \dnn{}s~\cite{Qin19Speech, Schon18AdvSound}. 

In response, researchers have proposed methods to mitigate the risks of
adversarial examples. For example, one method, called adversarial training,
augments the training data with correctly labeled adversarial examples 
(e.g.,~\cite{Kantchelian16ICML,
  Kurakin16AdvTrain, Madry17AdvTraining, Shafahi19PGD}). The resulting models
are often more robust in the face of attacks than models trained via standard
methods.\lujo{don't really need previous sentence}
However, while defenses are constantly improving, they are still far
from perfect. Relative to standard models, defenses often reduce the accuracy
on benign samples. For example, methods to detect the presence of attacks
sometimes erroneously detect benign inputs as adversarial (e.g.,~\cite{Ma18LID,
  Ma19Detection}). Moreover, defenses often fail to mitigate a large fraction of
adversarial examples that are produced by strong attacks
(e.g.,~\cite{Athalye18Attack}).

Inspired by \nvariants{}-version programming, this paper proposes a new defense,
termed \nML{}, that improves upon the state of the art in its ability
to detect adversarial inputs and correctly classify benign ones.
Similarly to other ensemble classifiers~\cite{Machida19nML, Strauss17Ensemble,
  Xu19nvDNN}, an \nML{} ensemble outputs 
the majority vote if more than a threshold number of \dnn{}s agree;
otherwise the input is deemed adversarial.  
The key innovation in this work is a novel method, \emph{topological
  manipulation},
to train \dnn{}s to achieve high
accuracy on benign samples while simultaneously classifying adversarial
examples \emph{according to specifications that are drawn at random before training}.
Because every \dnn{} in an ensemble is trained to classify adversarial
examples differently than the other \dnn{}s, \nML{} is able to detect
adversarial examples because they cause disagreement between the \dnn{}s'
votes.

We evaluate \nML{} using three datasets (\mnist{}~\cite{MNIST},
\cifarten~\cite{Krizhevsky09CIFAR}, and \gtsrb{}~\cite{Stallkamp12GTSRB})
and against \lpnorm{\infty} (mainly) and \lpnorm{2} attacks in black-,
grey-, and white-box settings. Our findings indicate that \nML{} can
effectively mitigate adversarial examples while achieving high benign
accuracy. For example, for \cifarten{} in the black-box setting, \nML{} can
achieve 94.50\% benign accuracy (vs.\ 95.38\% for the best standard \dnn{}) while
preventing all adversarial examples with \lpnorm{\infty}-norm perturbation
magnitudes of $\frac{8}{255}$ created by the best known attack
algorithms~\cite{Madry17AdvTraining}.
\lujo{is $\infty$-norms being used as a unit? is some preposition needed?}\mahmood{fixed}
In comparison, the state-of-the-art 
defense achieves 87.24\%
benign accuracy while being evaded by 14.02\% of the adversarial
examples. \nML{} is also faster than most defenses that we compare against.
Specifically, even the slowest variant of \nML{} is $\times$45.72 to
$\times$199.46 faster at making inferences than other defenses for detecting
the presence of attacks.\lujo{would be nice to qualify ``fast''. inference time?
  some other time?}\mahmood{addressed (but forgot to mention when I committed first :-))}

Our contributions can be summarized as follows:\footnote{We will release our
  implementation of \nML{} upon publication.}
\begin{itemize}
  \item We propose \emph{topology manipulation}, a novel method to train \dnn{}s
    to classify adversarial examples according to specifications
    that are selected at training time, while also achieving high benign
    accuracy.
  \item Using topologically manipulated \dnn{}s, we construct (\nML{}) ensembles to
    defend against adversarial examples.
  \item Our experiments using two perturbation types and three
    datasets in black-, grey-, and white-box settings show that \nML{}
    is an effective and efficient defense. \nML{} 
    roughly retains the benign accuracies of state-of-the-art \dnn{}s, while
    providing more
    resilience to attacks than the best defenses known to date, and making
    inferences faster than most.
\end{itemize}

We next present the background and related work (\secref{sec:relwork}). Then, we
present the technical details behind \nML{} and topology manipulation
(\secref{sec:method}). Thereafter, we describe the experiments that we conducted
and their results (\secref{sec:res}). We close the paper with a discussion
(\secref{sec:discuss}) and a conclusion (\secref{sec:conclude}).

\section{Background and Related Work}
\label{sec:relwork}

This section summarizes related work and provides the necessary background on
evasion attacks on \ml{} algorithms and defenses against them.

\subsection{Evading \ml{} Algorithms}

Inputs that are minimally perturbed to fool \ml{} algorithms at inference
time---termed adversarial examples---have emerged as a challenge to \ml{}. 
Attacks to produce adversarial examples typically start from benign inputs and
find perturbations with small \lpnorm{p}-norms ($p$ typically
$\in\{0,2,\infty\}$) that lead to misclassification when added to the benign
inputs (e.g.,~\cite{Baluja17ATNs, Biggio13Evasion, Carlini17Robustness,
  Gdfllw14ExpAdv, Moosavi16DeepFool, Papernot16Limitations, Szegedy13NNsProps,
  Xiao18GANAttack}). Keeping the perturbations' \lpnorm{p}-norms small helps
ensure the attacks' imperceptibility to humans, albeit imperfectly~\cite{Sen19Lp,
  Sharif18Lp}. The process of finding adversarial
perturbations is usually formalized as an optimization problem. For example,
Carlini and  Wagner~\cite{Carlini17Robustness} proposed the following
formulation to find adversarial perturbations that target class $c_t$:
$$\arg\min_{r}\  \cwloss(x+r,c_t) + \kappa\cdot||r||_{2}$$ where $x$ is a benign
input, $r$ is the perturbation, and $\kappa$ helps tune the \lpnorm{2}-norm of
the perturbation. $\cwloss$ is roughly defined as:
$$  \cwlosstarg(x+r,c_t) = \max_{c\neq c_t} \{ \dnnLogits_{c}(x+r) \} - \dnnLogits_{c_t}(x+r)$$
where $\dnnLogits_{c}$ is the output for class $c$ at the logits of the
\dnn{}---the output of the one-before-last layer. Minimizing $\cwloss$  leads
$x+r$ to be (mis)classified as $c_t$. As this formulation targets a specific
class $c_t$, the resulting attack is commonly referred to as a \emph{targeted
  attack}. In the case of evasion where the aim is to produce an adversarial
example that is misclassified as any class but the true class (commonly referred
to as \emph{untargeted attack}), $\cwloss$  is defined as:
$$  \cwlossuntarg(x+r,c_x) = \dnnLogits_{c_x}(x+r) - \max_{c\neq c_x} \{ \dnnLogits_{c}(x+r) \}$$
where $c_x$ is the true class of $x$. We use $\cwloss$ as a loss function to fool
\dnn{}s.

The Projected Gradient Descent (\pgd{}) attack is considered as the strongest
first-order attack (i.e., an attack that uses gradient decent to find
adversarial examples)~\cite{Madry17AdvTraining}. Starting from any random point
close to a benign input, \pgd{} consistently finds perturbations with
constrained \lpnorm{\infty}- or \lpnorm{2}-norms that achieve roughly the same
loss value. Given a benign input $x$, a loss function, say $\cwloss$, a target
class $c_t$, a step size $\alpha$, and an upper bound on the norm $\epsilon{}$, \pgd{}
iteratively updates the adversarial example until a maximum number of iterations
is reached such that:  
$$x_{i+1} = \mathit{Project}_{x,\epsilon}(x_{i}+\alpha{}\mathit{sign}(\nabla_{x}\cwloss(x_{i}, c_t))) $$
where $\mathit{Project}_{x,\epsilon}(\cdot)$ projects vectors on the
$\epsilon$-ball around $x$ (e.g., by clipping for \lpnorm{\infty}-norm), and
$\nabla_{x}\cwloss(\cdot)$ denotes the gradient of the loss function. \pgd{}
starts from a random point $x_0$ in the  $\epsilon$-ball around $x$, and stops
after a fixed number of iterations (20--100 iterations are
typical~\cite{Madry17AdvTraining, Shafahi19PGD}). In this
work, we rely on \pgd{} to produce adversarial examples.

Early on, researchers noticed that adversarial examples computed against one \ml{}
model are likely to be misclassified by other models performing the same
task~\cite{Gdfllw14ExpAdv, Papernot17Blackbox, Papernot16Transferability}. This
phenomenon---transferability---serves as the basis of
non-interactive attacks in black-box settings, where attackers do not have
access to the attacked models and may not be able to query them. The accepted
explanation for the transferability phenomenon, at least between \dnn{}s
performing the same task, is that the gradients (which are used to compute
adversarial examples) of each \dnn{} are good approximators of those of other
\dnn{}s~\cite{Demontis19Transfer, Ilyas19Explain}. Several techniques can be
used to enhance the transferability of adversarial examples~\cite{Dong19TI,
  Liu17Transfer, Tramer18Ensemble, Xie19ID}. The techniques include computing
adversarial perturbations against ensembles of \dnn{}s~\cite{Liu17Transfer,
  Tramer18Ensemble} and misleading the \dnn{}s after transforming the input (e.g.,
by translation or resizing)~\cite{Dong19TI, Xie19ID} as a means to avoid
computing perturbations that generalize beyond a single \dnn{}. We leverage
these techniques to create strong attacks against which we evaluate
our proposed defense. 

\subsection{Defending \ml{} Algorithms}

Researchers proposed a variety of defenses to mitigate \ml{} algorithms'
vulnerability to adversarial examples. Roughly speaking, defenses can be
categorized as: 
adversarial training, certified defenses, attack
detection, or input transformation. Similarly to our proposed defense, some
defenses leverage randomness or model ensembles. In what follows, we provide an
overview of the different categories.

\parheading{Adversarial Training}
Augmenting the training data with correctly labeled adversarial examples that
are generated throughout the training process increases models' robustness to
attacks~\cite{Gdfllw14ExpAdv, Kantchelian16ICML, Kannan18ALP, Kurakin16AdvTrain,
  Madry17AdvTraining, Shafahi19PGD, Szegedy13NNsProps}. The resulting training
process is commonly referred to as \emph{adversarial training}. In particular,
adversarial training with \pgd{} (\advpgd{})~\cite{Madry17AdvTraining, Shafahi19PGD} is one
of the most effective defenses to date---we compare our defense against it. 

\parheading{Certified Defenses}
Some defenses attempt to certify the robustness
of trained \ml{} models (i.e., provide provable bounds on models' errors for
different perturbation magnitudes). Certain certified defenses estimate how
\dnn{}s transform \lpnorm{p} balls around benign examples via convex shapes,
and attempt to force classification boundaries to not cross the shapes
(e.g.,~\cite{Kolter17Defense, Mirman18Defense, Raghu18Certified}). These
defenses are less effective than adversarial training with
\pgd{}~\cite{Salman19Convex}. Other defenses estimate the output of
the so-called \textit{smoothed classifier} by classifying many
variants of the input 
after adding noise at the input or intermediate layers~\cite{Cohen19RandSmooth,
  Lecuyer19PixelDP, Liu18RSE, Salman19Robust}. The resulting smoothed
classifier, in turn, is proven to be robust against perturbations of different
\lpnorm{2}-norms. Unfortunately, such defenses do not provide guarantees against
perturbations of different types (e.g., ones with bounded
\lpnorm{\infty}-norms), and perform less well against them in
practice~\cite{Cohen19RandSmooth, Lecuyer19PixelDP}.

\parheading{Attack Detection}
Similarly to our defense, there were past proposals for defenses to detect the
presence of attacks~\cite{Abbasi17Ensemble, Feinman17Detector, Grosse17Detector, Huang19Random,
  Lu18FoolLID, Ma19Detection, Meng17Magnet, Metzen17Detector, Pang18DetectRCE,
  Wang19Mutation}. While adaptive attacks have been shown to circumvent some of
these defenses~\cite{Athalye18Attack, Carlini17Bypass, Lu18FoolLID}, detectors
often significantly increase the magnitude of the perturbations necessary to
evade \dnn{}s and detectors combined~\cite{Carlini17Bypass, Ma19Detection}. In
this work, we compare our proposed defense with detection methods based on Local Intrinsic
Dimensionality (\lid{})~\cite{Ma18LID} and Network Invariant Checking
(\nic{})~\cite{Ma19Detection}, which are currently the leading methods for
detecting adversarial examples.

The \lid{} detector uses a logistic regression classifier to tell benign and
adversarial inputs apart. The input to the classifier is a vector of \lid{}
statistics that are estimated for every intermediate representation computed by
the \dnn{}. This approach is effective because the \lid{} statistics
of adversarial examples are presumably distributed differently than those of
benign inputs~\cite{Ma18LID}.

The \nic{} detector expects certain invariants in \dnn{}s to hold for benign
inputs. For example, it expects the provenance of activations for benign inputs to follow a certain
distribution. To model these invariants, a linear logistic
regression classifier performing the same task as the original \dnn{} is trained
using the representation of every intermediate layer. Then, for every pair of
neighboring layers, a one-class Support Vector Machine (\ocsvm{}) is trained to model the
distribution of the output of the layers' classifiers on benign inputs.
Namely, every \ocsvm{} receives concatenated vectors of probability estimates
and emits a score indicative of how similar the vectors are to the benign distribution.
The scores of all the \ocsvm{}s are eventually combined to derive an estimate for
whether the input is adversarial. In this manner, if the output of two
neighboring classifiers on an image, say that of a bird, is $(\textit{bat},\textit{bird})$,
the input is likely to be benign (as the two classes are similar and likely have
been observed for benign inputs during training). However, if
the output is $(\textit{car}, \textit{bird})$,  then it is likely that the
input is adversarial.

\parheading{Input Transformation}
Certain defenses suggest to transform inputs (e.g., via quantization) to
sanitize adversarial perturbations before classification~\cite{Guo18ReformDef,
  Liao18ReformDef, Meng17Magnet, Samangouei18DefGAN, Srini18ReformDef,
  Xie18Denoise, Xu18Squeeze}. The transformations often aim to hinder the
process of computing gradients for the purpose of attacks. In practice, however,
it has been shown that attackers can adapt to circumvent such
defenses~\cite{Athalye18BeatCVPR, Athalye18Attack, He17FoolEnsembles}.

\parheading{Randomness}
Defenses often leverage randomness to mitigate adversarial examples. As
previously mentioned, some defenses inject noise at inference time at the input
or intermediate layers~\cite{Cohen19RandSmooth, Huang19Random, Lecuyer19PixelDP,
  Liu18RSE, Salman19Robust}. Differently, Wang et al.\ train a hierarchy of
layers, each containing multiple paths, and randomly switch between the chosen
paths at inference time~\cite{Wang19HRS}. \lujo{not clear what ``blocks''
  are}\mahmood{removed ``blocks''}
Other defenses randomly drop out neurons, shuffle them, or change their
weights, also while making inferences~\cite{Feinman17Detector,
  Wang19Mutation}. Differently from all 
these, our proposed defense uses randomness at \emph{training time} to control how a
set of \dnn{}s classify adversarial examples at inference time and uses the
\dnn{}s strategically in an ensemble to deter adversarial examples.
\lujo{too weak, makes it sound like we use only randomness}\mahmood{extended to
  say more}

\parheading{Ensembles}
Similarly to our defense, several prior defenses suggested using ensembles to
defend against adversarial examples. Abbasi and Gagn\'e proposed to measure the
disagreement between \dnn{}s and specialized classifiers (i.e., ones that
classify one class versus all others) to detect adversarial
examples~\cite{Abbasi17Ensemble}. An adaptive attack to evade the specialized
classifiers and the \dnn{}s simultaneously can circumvent this
defense~\cite{He17FoolEnsembles}. Vijaykeerthy et al.\ train \dnn{}s
sequentially to be robust against an increasing set of
attacks~\cite{Vijay19Cascades}. However, they only use the final model at
inference time, while we use an ensemble containing several models for
inference. 
A meta defense by Sengupta et
al.\ strategically selects a model from a pool of candidates at inference time
to increase benign accuracy while deterring
attacks~\cite{Sengupta19MTDeep}. This defense is effective against black-box 
attacks only. Other ensemble defenses propose novel training or inference
mechanisms, but do not achieve competitive performance~\cite{Kari19EnsembleDiv,
  Pang19Ensembles, Strauss17Ensemble}.
\lujo{anything positive to be said?}\mahmood{done}

Recent papers~\cite{Machida19nML, Xu19nvDNN, Zeng19MVP} proposed defenses that,
similarly to ours, are motivated by \nvariants-version programming~\cite{Avizienis85nVP,
  Chen95nVP, Cox06nVS}. In a nutshell, \nvariants-version programming aims to provide
resilience to bugs and attacks by running $\nvariants$ ($\ge$2) variants
of independently developed, or diversified, programs. These programs are
expected to behave identically for normal (benign) inputs, and differently for
unexpected inputs that trigger bugs or exploit vulnerabilities. When one or more
programs behaves differently than the others, an unexpected (potentially
malicious) input is detected. In the context of \ml{}, defenses that are
inspired by \nvariants-version programming use ensembles of models that are developed by
independent parties~\cite{Zeng19MVP}, different inference algorithms or \dnn{}
architectures~\cite{Machida19nML, Xu19nvDNN},  or models that are trained using
different training sets~\cite{Xu19nvDNN}. In all cases, the models are
trained via standard training techniques. Consequently, the 
defenses are often vulnerable to attacks (both in black-box and more challenging
settings) as adversarial examples transfer with high likelihood regardless of
the inference algorithm or the training data~\cite{Gdfllw14ExpAdv,
  Papernot17Blackbox, Papernot16Transferability}. Moreover, prior work is
limited to specific applications  (e.g., speech recognition~\cite{Zeng19MVP}).
In contrast, we train the models comprising the ensembles via a novel training
technique (see \secref{sec:method}), and our work is conceptually applicable to
any domain.

\section{Technical Approach}
\label{sec:method}

Here we detail our methodology. We begin by presenting the threat
model. Subsequently, we present a novel technique, termed \emph{topology manipulation},
which serves as a cornerstone for training \dnn{}s that are used as part of the
\nML{} defense. Last, we describe how to construct an \nML{} defense via an
ensemble of topologically manipulated \dnn{}s.

\subsection{Threat Model}

Our proposed defense aims to mitigate attacks in \mbox{\emph{black-},}
\emph{grey-}, and \emph{white-box} settings. In the black-box setting, the
attacker has no access to the classifier's parameters and is unaware of the
existence of the defense.\footnote{For clarity, we distinguish the classifier
  from the defense. However, in certain cases (e.g., for \nML{} or
  adversarial training) they are inherently inseparable.} To evade
classification, the attacker attempts a non-interactive attack by transferring
adversarial examples from standard surrogate models. Similarly, in the grey-box
setting, the attacker cannot access the classifier's parameters. However, the
attacker is aware of the use of the defense and attempts to
transfer adversarial examples that are produced against a surrogate defended
model. In the white-box setting, the attacker has complete access to the
classifier's and defense's parameters. Consequently, the attacker can adapt 
gradient-based white-box attacks (e.g., \pgd{}) to evade classification. 
We do not consider interactive attacks that query models in a black-box setting
(e.g.,~\cite{Brendel18DBA, Brunner18BBox, Chen17Zoo, Ilyas19Explain}). These
attacks are generally weaker than the white-box attacks that we do consider.

As is typical in the area (e.g.,~\cite{Kolter17Defense, Madry17AdvTraining,
  Salman19Convex}), we focus on defending against adversarial perturbations with
bounded \lpnorm{p}-norms. In particular, we mainly consider defending against
perturbations with bounded \lpnorm{\infty}-norms. Additionally, we demonstrate
defenses against perturbations with bounded \lpnorm{2}-norms as a proof of
concept. Conceptually, there is no reason why \nML{} should not
generalize to defend against other types of attacks.

\subsection{Topologically Manipulating \dnn{}s}

\begin{figure}
  \centering
  \includegraphics[width=0.48\columnwidth]{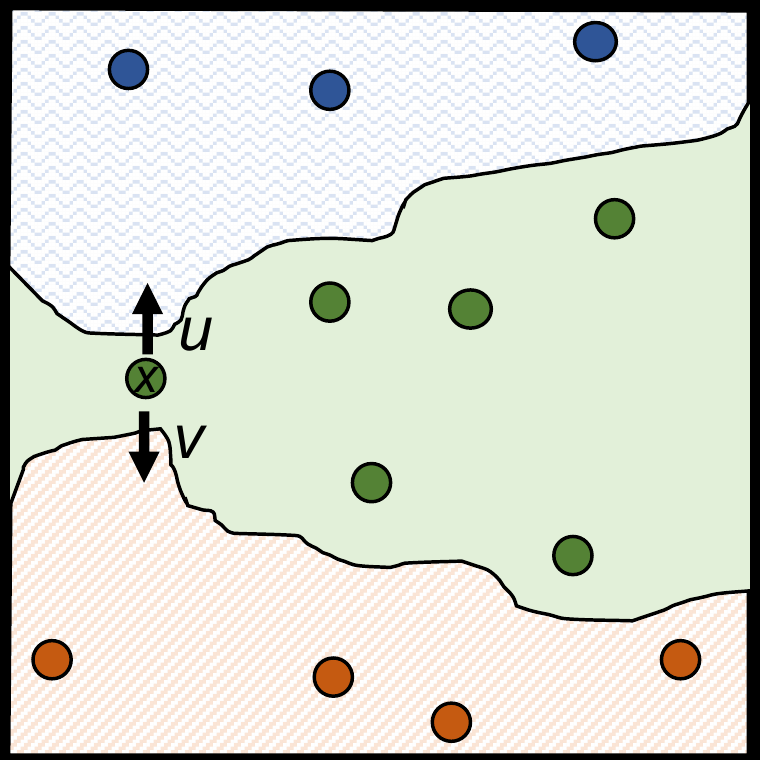}
  \includegraphics[width=0.48\columnwidth]{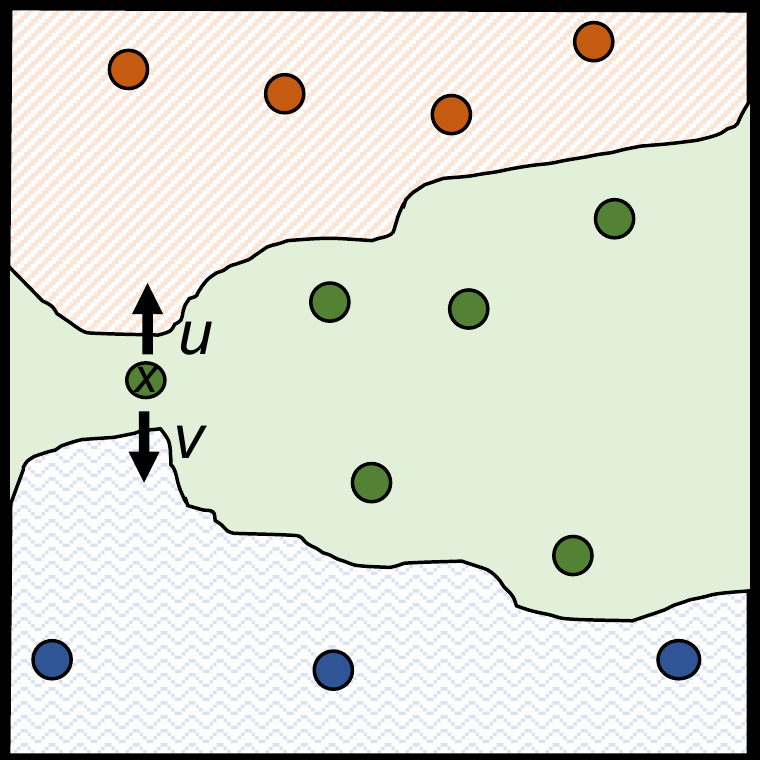}
  \caption{\label{fig:illus_abs} An illustration of topology manipulation. Left: In a
    standard \dnn{}, perturbing the benign sample $x$ in the direction of $u$
    leads to misclassification as blue (zigzag pattern), while perturbing it in
    the direction of $v$ leads to misclassification as red (diagonal stripes).
    Right: In the topologically manipulated \dnn{}, direction $u$ leads to
    misclassification as red, while $v$ leads to misclassification as blue. The
    benign samples (including $x$) are correctly classified in both cases (i.e.,
    high benign accuracy).}
\end{figure}

\begin{figure}
  \centering
  \begin{subfigure}[h]{0.16\columnwidth}
    \centering
    \includegraphics[width=\textwidth]{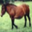}
    \caption{}
  \end{subfigure}\hspace{1pt}
  \begin{subfigure}[h]{0.4\columnwidth}
    \centering
    \includegraphics[width=0.4\textwidth]{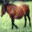}
    \includegraphics[width=0.4\textwidth]{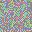}
    \caption{}
  \end{subfigure}\hspace{1pt}
  \begin{subfigure}[h]{0.4\columnwidth}
    \centering
    \includegraphics[width=0.4\textwidth]{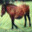}
    \includegraphics[width=0.4\textwidth]{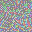}
    \caption{}
  \end{subfigure}
  \caption{\label{fig:illus_con}A concrete example of topology manipulation. The
  original image of a horse (a) is adversarially perturbed to be misclassified
  as a bird (b) and as a ship (c) by standard \dnn{}s. The perturbations, which
  are limited to \lpnorm{\infty}-norm of ${8 \over 255}$, are shown after
  multiplying $\times$10. We train a topologically manipulated \dnn{} to
  misclassify (b) as a ship and (c) as a bird, while classifying the original
  image correctly.}
\end{figure}

The main building block of the \nML{} defense is a \emph{topologically
  manipulated \dnn{}}---a \dnn{} that is manipulated at training time to achieve
certain topological properties with respect to adversarial
examples. Specifically, a topologically manipulated \dnn{} is trained to satisfy
two objectives:
\emph{1)} obtaining high classification accuracy on benign inputs;
and \emph{2)} misclassifying adversarial inputs following a certain specification.
The first objective is important for constructing a well-performing \dnn{} to
solve the classification task at hand. The second objective aims to
change the adversarial directions of the \dnn{} such that an adversarial
perturbation that would normally lead a benign input to be misclassified as
class $c_t$ by a regularly trained \dnn{} would likely lead to misclassification as
class $\hat{c_t}$ ($\neq c_t$) by the topologically manipulated \dnn{}.
\figref{fig:illus_abs} illustrates the idea of manipulating the topology of a
\dnn{} via an abstract example, while \figref{fig:illus_con} gives a concrete
example.

To train a topologically manipulated \dnn{}, two datasets are used. The
first dataset, $D$, is a standard dataset. It contains pairs $(x, c_x)$ of
benign samples, $x$, and their true classes, $c_x$. The second dataset,
$\tilde{D}$, contains adversarial examples. Specifically, it consists of pairs
$(\tilde{x}, c_t)$ of targeted adversarial examples, $\tilde{x}$, and the
target classes, $c_t$. These adversarial examples are produced against
reference \dnn{}s that are trained in a standard manner (e.g., to decrease
cross-entropy loss). Samples in $D$ are used to train the \dnn{}s to satisfy
the first objective (i.e., achieving high benign accuracy). Samples in
$\tilde{D}$, on the other hand, are used to topologically manipulate the
adversarial directions of \dnn{}s.

More specifically, to specify the
topology of the trained \dnn{}, we use a derangement (i.e., a permutation with
no fixed points), $d$, that is drawn at random over the number of classes,
$\nclasses$. This derangement specifies that an adversarial example $\tilde{x}$ in
$\tilde{D}$ that targets class $c_t$ should be misclassified as $d[c_t]$ ($\neq c_t$) by the
topologically manipulated \dnn{}. For example, for ten classes (i.e., $\nclasses=10$),
the derangement may look like $d=[1, 6, 7, 0, 2, 8, 9, 3, 5, 4]$. This
derangement specifies that adversarial examples targeting class 0 should be
misclassified as class $d[0]=1$, ones targeting class 1 should be
misclassified as class $d[1]=6$, and so on. For $\nclasses{}$ classes, the number of
derangements the we can draw from is known as the subfactorial (denoted by
$!\nclasses{}$), and is defined recursively as
$!\nclasses{}=(\nclasses{}-1)(!(\nclasses{}-1)+!(\nclasses{}-2))$, where $!2=1$ and
$!1=0$.
The subfactorial grows almost as quickly as the factorial $\nclasses{}!$
(i.e., the number of permutations over a group of size $\nclasses{}$).

We specify the topology using derangements rather than permutations that may
have fixed points because
if $d$ contained fixed points, there would exist a class
$c_t$ such that $d[c_t]=c_t$. In such case, the \dnn{} would be trained to
misclassify adversarial examples that target $c_t$ into $c_t$, which would not inhibit an adversary targeting $c_t$. Such behavior is undesirable.

We use the standard cross-entropy loss, $\celoss{}$,\footnote{In the case of one-hot
encoding for the true class $c_x$ and a probability estimate $\hat{p}_{c_x}$ emitted by
the \dnn{}, the cross-entropy loss is defined as  $-\log(\hat{p}_{c_x})$.} 
to train topologically manipulated \dnn{}s.
Formally, the training process minimizes:
\begin{equation}
  \frac{1}{|D|} \sum_{(x, c_{x})\in D} \celoss(x, c_{x}) + \frac{\lambda}{|\tilde{D}|} \sum_{(\tilde{x}, c_{t})\in \tilde{D}} \celoss(\tilde{x}, d[c_{t}])
\label{eqn:nmlobj}
\end{equation}
While minimizing the leftmost term increases the benign accuracy (as is usual in
standard training processes), minimizing the rightmost term manipulates the
topology of the \dnn{} (i.e., forces the \dnn{} to misclassify $\tilde{x}$ as
$d[c_t]$ instead of as $c_t$). The parameter $\lambda$ is a positive real number
that balances the two objectives. We tune it via a hyperparameter search.

Although topologically manipulated \dnn{}s aim to satisfy multiple objectives,
it is important to point out that training them does not require significantly
more time than for standard \dnn{}s. For training, one first needs to create the
dataset $\tilde{D}$ that contains the adversarial examples. This needs to be
done only once, as a preprocessing phase. Once $\tilde{D}$ is created, training
a topologically manipulated \dnn{} takes the same amount of time as training a
standard \dnn{}.

\subsection{\nML{}: An Ensemble-Based Defense}

As previously mentioned, \nML{} is inspired by $\nvariants$-version programming. While
$\nvariants{}$ independent, or diversified, programs are used in an $\nvariants$-version
programming defense, an \nML{} defense contains an ensemble of $\nvariants{}$
($\ge$2) topologically manipulated \dnn{}s. As explained above, all the \dnn{}s
in the \nML{} ensemble are trained to behave identically for benign inputs
(i.e., to classify them correctly), while each \dnn{} is trained to 
follow a different specification for adversarial examples. This opens an
opportunity to \emph{1)} classify benign inputs accurately; and \emph{2)}
detect adversarial examples.

In particular, to classify an input $x$ using an \nML{} ensemble, we compute the
output of all the \dnn{}s in the ensemble on $x$. Then, if the number of \dnn{}s
that agree on a class is above or equal to a threshold $\threshold$, the input is
classified to the majority class. Otherwise, the \nML{} ensemble would abstain
from classification and the input would be marked as adversarial. Formally,
denoting the individual \dnn{}s' classification results by
the multiset $C=\{\dnnFunc{i}(x)| 1\le i\le\nvariants{}\}$,
the \nML{} classification function, $\nMLFunc$, is defined as:
\[
\nMLFunc(x)=
\begin{cases}
  \mathit{majority}(C), & \text{if}~|\{c\in C| c=\mathit{majority}(C)\}|\ge\threshold  \\
  \mathit{abstain}, & \text{otherwise}
\end{cases}
\]

Of course, increasing the threshold increases the likelihood of
detecting adversarial examples (e.g., an adversarial example is less likely to
be misclassified as the same target class $c_t$ by all the $\nvariants$ \dnn{}s than by
$\nvariants{}-1$ \dnn{}s). In other words, increasing $\threshold$ decreases attacks'
success rates. At the same time, increasing the threshold harms the 
benign accuracy (e.g., the likelihood of $\nvariants$ \dnn{}s to emit $c_x$ is
lower than the likelihood of $\nvariants{}-1$ \dnn{}s to do so). In practice, we
set $\threshold$ to a value $\ge\lceil{\frac{\nvariants{}+1}{2}}\rceil$, to avoid
ambiguity when computing the majority vote, and $\le\nvariants{}$, as the benign
accuracy is 0 for $\threshold >\nvariants$.\lujo{this last seems a little obvious}

Similarly to $\nvariants$-version programming, where the defense becomes more effective
when the constituent programs are more independent and
diverse~\cite{Avizienis85nVP, Chen95nVP, Cox06nVS}, an \nML{} defense is more
effective at detecting adversarial examples when the \dnn{}s are more
independent. Specifically, if two \dnn{}s $i$ and $j$ ($i\neq{}j$) are trained
with derangements $d_i$ and $d_j$, respectively, and we are not careful enough,
there might exist a class $c_t$ such that $d_i[c_t]=d_j[c_t]$. If so, the two
\dnn{}s are likely to classify adversarial examples targeting $c_t$ in the
same manner, thus reducing the defense's likelihood to detect attacks. To
avoid such undesirable cases, we train the \nML{} \dnn{}s (simultaneously or
sequentially) while attempting to avoid pairs of derangements that map
classes in the same manner to the greatest extent possible. More concretely, if
$\nvariants{}$ is lower than the number of classes $\nclasses{}$, then we draw
$\nvariants{}$ derangements that disagree on all indices (i.e.,
$\forall{}i\neq{}j$, $\forall{}c_t$, $d_i[c_t]\neq{}d_j[c_t]$). Otherwise, we split 
the \dnn{}s to groups of $\nclasses{}-1$ (or smaller) \dnn{}s, and for each
group we draw derangements that disagree on all indices. For a group of
$\nvariants{}$ \dnn{}s, $\nvariants{}<\nclasses{}$, we can draw
$\prod_{i=1}^{\nvariants{}} !(\nclasses{}-i+1)$ derangements such that every pair
of derangements disagrees on all indices.

\section{Results}
\label{sec:res}

In this section we describe the experiments that we conducted and their
results. We initially present the datasets and the standard \dnn{}
architectures that we used. Then we describe how we trained individual topologically
manipulated \dnn{}s to construct \nML{} ensembles, and evaluate the extent to 
which they met the training objectives. We close the section with experiments
to evaluate the \nML{} defense in various settings. We ran our experiments with
Keras~\cite{Keras15} and TensorFlow~\cite{Abadi16TensorFlow}.

\subsection{Datasets}

We used three popular datasets to evaluate \nML{} and other defenses:
\mnist{}~\cite{MNIST}, \cifarten{}~\cite{Krizhevsky09CIFAR}, and
\gtsrb~\cite{Stallkamp12GTSRB}. \mnist{} is a dataset of $28\times{}28$ pixel
images of digits (i.e., ten classes). It contains 70,000 images in total, with
60,000 images intended for training and 10,000 intended for testing. We
set aside 5,000 images from the training\lujo{testing? this means we split
  the test set into two, one of which we don't use for testing?}\mahmood{right,
  but from the training set.} set for
validation. \cifarten{} is
a dataset of $32\times{}32$-pixel images of ten classes: airplanes, automobiles,
birds, cats, deer, dogs, frogs, horses, ships, and trucks. The dataset contains
50,000 images for training and 10,000 for testing. We set aside 5,000 images
from the training set for validation. Last, \gtsrb{} is a dataset
containing traffic signs of 43 classes. The dataset contains 39,209 training
images and 12,630 test images. We used 1,960 images that we set aside from
the training set for validation.\lujo{i'm confused now about from
where we draw validation images and why it's different for different
datasets. do we use them in training if they're drawn from the
training set? if not, can we say we ``set aside'' instead of
``draw''?}\mahmood{fixed. agreed, ``set aside'' is better.}
Images in \gtsrb{} vary in size between
$15\times{}15$ and $250\times{}250$. Following prior work~\cite{Li18GTSRB}, we
resized the images to $48\times{}48$.

The three datasets have different properties that made them especially suitable
for evaluating \nML{}. \mnist{} has relatively few classes, thus limiting the
set of derangements that we could use for topology manipulation for large values of \nvariants{}. At the same time, standard \dnn{}s achieve high classification accuracy
on \mnist{} ($\sim$99\% accuracies are common), hence increasing the likelihood
that \dnn{}s in the ensemble would agree on the correct class for benign inputs.
\cifarten{} also relatively has a few classes. However, differently from
\mnist{}, even the best performing \dnn{}s do not surpass $\sim$95\%
classification accuracy on \cifarten{}. Consequently, the likelihood that a large
number $\nvariants$ of \dnn{}s in an ensemble would achieve consensus may be low (e.g., an
ensemble consisting of $\nvariants = 5$ \dnn{}s with 95\% accuracy each that incur independent
errors could have benign accuracy as low as 77\% if we require all \dnn{}s to agree). 
In contrast, \gtsrb{}
contains a relatively high number of classes, and standard \dnn{}s often achieve
high classification accuracies on this dataset (98\%--99\% accuracies are
common). As a result, there is a large space from which we could draw derangements
for topology manipulation, and we had expected high benign accuracies from \nML{}
ensembles.

\subsection{Training Standard \dnn{}s}

\begin{table}[ht]
  \centering
  \begin{tabular}{c | c | l | c}
    \toprule
    \emph{Dataset} & \# & \emph{Architecture} & \emph{Acc.} \\ \midrule
    \multirow{6}{*}{\mnist{}}
    & 1 & Convolutional \dnn{}~\cite{Carlini17Robustness} & 99.42\% \\
    & 2 & Convolutional \dnn{}~\cite{Madry17AdvTraining} & 99.28\% \\
    & 3 & Convolutional \dnn{}~\cite{Carlini17Robustness} w/o pooling~\cite{Spring15AllConv} & 99.20\% \\
    & 4 & Convolutional \dnn{}~\cite{KerasMNISTCNN} & 99.10\% \\
    & 5 & Convolutional \dnn{}~\cite{Madry17AdvTraining} w/o pooling~\cite{Spring15AllConv} & 99.10\% \\
    & 6 & Multi-layer perceptron~\cite{KerasMNISTMLP} & 98.56\% \\
    \midrule

    \multirow{6}{*}{\cifarten{}}
    & 1 & \wresnet{22}{8}~\cite{Zagoruyko16WRN} & 95.38\% \\
    & 2 & \wresnet{28}{10}~\cite{Zagoruyko16WRN} & 95.18\% \\
    & 3 & \wresnet{16}{10}~\cite{Zagoruyko16WRN} & 95.06\% \\
    & 4 & \wresnet{28}{8}~\cite{Zagoruyko16WRN} & 94.88\% \\
    & 5 & \wresnet{22}{10}~\cite{Zagoruyko16WRN} & 94.78\% \\
    & 6 & \wresnet{16}{8}~\cite{Zagoruyko16WRN} & 94.78\% \\
    \midrule

    \multirow{6}{*}{\gtsrb{}}
    & 1 & Convolutional \dnn{}~\cite{Li18GTSRB} & 99.46\% \\
    & 2 & Same as 1, but w/o first branch~\cite{Li18GTSRB} & 99.56\% \\
    & 3 & Same as 1, but w/o pooling~\cite{Spring15AllConv} & 99.11\% \\
    & 4 & Same as 1, but w/o second branch~\cite{Li18GTSRB} & 99.08\% \\
    & 5 & Convolutional \dnn{}~\cite{Tian17GTSRB} & 99.00\% \\
    & 6 & Convolutional \dnn{}~\cite{Sermanet11GTSRB} & 98.07\% \\
    
    \bottomrule
  \end{tabular}
  \caption{\label{tab:archs}The \dnn{} architectures that we used for the
    different datasets. The \dnn{}s' accuracies on the test sets of the
    corresponding datasets (after standard training) are reported to the right.}
\end{table}

The \dnn{}s that we used were based on standard architectures. We constructed the
\dnn{}s either exactly the same way as prior work or reputable public projects
(e.g., by the Keras team~\cite{KerasMNISTCNN}) or by modifying prior \dnn{}s via
a standard technique. In particular, we modified certain \dnn{}s following the
work of Springenberg et al.~\cite{Spring15AllConv}, who found that it is
possible to construct simple, yet highly performing, \dnn{}s by removing pooling
layers (e.g., max- and average-pooling) and increasing the strides of
convolutional operations.

For each dataset, we trained six \dnn{}s of different architectures---a
sufficient number of \dnn{}s to allow us to evaluate \nML{} and
perform attacks via transferability from surrogate
ensembles~\cite{Liu17Transfer}
while factoring out the effect of architectures (see \secref{sec:res:topoman}
and \secref{sec:res:nml}).
\lujo{why 6? what do we do with them later?}\mahmood{added a reason}
The \mnist{} \dnn{}s were trained for 20 epochs using
the Adadelta optimizer with standard parameters and a batch size of
128~\cite{KerasMNISTCNN, Zeiler12Adadelta}. The \cifarten{} \dnn{}s were trained
for 200 epochs with data augmentation (e.g., image rotation and flipping) and
training hyperparameters set identically to prior work~\cite{Zagoruyko16WRN}.
The \gtsrb{} \dnn{}s were trained with the Adam optimizer~\cite{Kingma14Adam},
with training hyperparameters and augmentation following prior
work~\cite{Li18GTSRB, Sermanet11GTSRB, Tian17GTSRB}. \tabref{tab:archs} reports
the architectures and performance of the \dnn{}s. In all cases, the \dnn{}s
achieved comparable performance to prior work.

\subsection{Training Individual Topologically Manipulated \dnn{}s}
\label{sec:res:topoman}

Now we describe how we trained individual topologically manipulated
\dnn{}s and report on their performance. Later, in \secref{sec:res:nml},
we report on the performance on the \nML{} ensembles.

\parheading{Training} \lujo{for clarity, turn this into /textbf? doesn't help at-a-glance parsing as is}\mahmood{done}
When training topologically manipulated \dnn{}s, we aimed to minimize the loss
described in \eqnref{eqn:nmlobj}. To this end, we modified the
training procedures that we use to train the standard \dnn{}s in three ways:
\lujo{is it clear what ``standard'' is? instead add ``reported
in prior work'' after ``procedure''?}\mahmood{rephrased}
\begin{enumerate}
  \item We extended each batch of benign inputs with the same number of
    adversarial samples $(\tilde{x}, c_t)\in\tilde{D}$ and specified that
    $\tilde{x}$ should be classified as $d[c_t]$.
  \item In certain cases, we slightly increased the number of training epochs
    to improve the performance of the \dnn{}s.
  \item We avoided data augmentation for \gtsrb{}, as we found that it harmed the
    accuracy of (topologically manipulated) \dnn{}s on benign inputs.
\end{enumerate}

To set $\lambda$ (the parameter that balances the \dnn{}'s benign accuracy and
the success of topology manipulation, see \eqnref{eqn:nmlobj}), we performed a
hyperparameter search. We experimented with values in $\{0.1, 0.5, 1, 2, 10\}$
to find 
the best trade-off between the \nML{} ensembles'
benign accuracy and their ability to mitigate attacks. We found that $\lambda=2$
achieved the highest accuracies at low attack success
rates.\lujo{criteria for ``best'' tradeoff?}\mahmood{done}

To train the best-performing \nML{} ensemble, one should select the best
performing \dnn{} architectures to train topologically manipulated
\dnn{}s. However, since the goal of this paper is to evaluate a defense, we aimed
to give the attacker the advantage to assess the resilience of the defense in a
worst-case scenario (e.g., so that the attacker could use the better held-out
\dnn{}s as surrogates in transferability-based attacks\lujo{be more precise. does
  the attacker use the better or worse dnn as a proxy for the real
  one? not obvious to me which one is an advantage to the attacker}\mahmood{fixed}).
Therefore, we selected the \dnn{}
architectures with the lower benign accuracy to train topologically manipulated
\dnn{}s. More specifically, for each dataset, we trained \nML{}
ensembles by selecting round robin from the architectures shown in
rows 4--6 from \tabref{tab:archs}.

Constructing a dataset of adversarial examples, $\tilde{D}$, is a key part of
training topologically manipulated \dnn{}s. As we aimed to defend against attacks
with bounded \lpnorm{\infty}-norms, we used the corresponding \pgd{} attack to
produce adversarial examples: For each training sample $x$ of class $c_x$, we
produced $\nclasses{}-1$ adversarial examples, one targeting every class
$c_t\neq{}c_x$. Moreover, we produced adversarial examples with perturbations of
different magnitudes to construct \nML{} ensembles that can resist attacks with varied
strengths. For \mnist{}, where attacks of magnitudes $\epsilon{}\le{}0.3$ are typically
considered~\cite{Madry17AdvTraining} we used $\epsilon\in{}\{0.1, 0.2, 0.3, 0.4\}$.
For \cifarten{} and \gtsrb{}, where attacks of magnitudes
$\epsilon{}\le{}\frac{8}{255}$ are typically considered~\cite{Madry17AdvTraining, Papernot18DkNN},
we used $\epsilon\in{}\{\frac{2}{255}, \frac{4}{255}, \frac{6}{255}, \frac{8}{255}\}$.
(Thus, in total, $|\tilde{D}|=4\times{}(\nclasses{}-1)\times{}|D|$.) We ran
\pgd{} for 40 iterations, since prior work found that his leads to successful
attacks~\cite{Madry17AdvTraining, Shafahi19PGD}.
Additionally, to avoid overfitting to the standard \dnn{}s that were used for training, 
we used state-of-the-art techniques to enhance the transferability
of the adversarial examples, both by making the examples invariant to spatial
transformations~\cite{Dong19TI, Xie19ID} and by producing them against
an ensemble of models~\cite{Liu17Transfer, Tramer18Ensemble}---three standard
\dnn{}s of architectures 4--6.

For each dataset, we trained a total of 18 topologically manipulated
\dnn{}s. Depending on the setting, we used a different subset of the \dnn{}s to
construct \nML{} ensembles (see \secref{sec:res:nml}). The \dnn{}s were split into two sets of
nine \dnn{}s each ($<\nclasses{}$ for \mnist{} and \cifarten{}), such that the
derangements of every pair of \dnn{}s in the same set disagreed on all
indices.\lujo{why split into two sets? why equal sizes?}\mahmood{rephrased}

\parheading{Evaluation}
Each topologically manipulated \dnn{} was trained with two objectives in mind:
classifying benign inputs correctly (i.e., high benign accuracy) and
classifying adversarial examples as specified by the derangement drawn at training
time. Here we evaluate the extent to which the \dnn{}s we trained met these
objectives. Note that these \dnn{}s were not meant to be used
individually, but instead in the ensembles evaluated in \secref{sec:res:nml}.
\lujo{not sure if this is needed, but i feel like readers may need
  reminding that this is all informational and not the main point yet}

\begin{table*}[t!]
  \centering
  \begin{tabular}{r | c c |c c c c c}
    \toprule
    & \multicolumn{2}{c |}{Standard} & \multicolumn{5}{c}{Topologically manipulated} \\
    \emph{Dataset} & \emph{Acc.} & \tsr{}
          &\emph{Acc.} & \tsr{} & \tsr{} \emph{h/o}
          & \msr{}  & \msr{} \emph{h/o}   \\ \midrule
    \mnist{}    & 99.30\%$\pm$0.09\% & 43.05\%$\pm$7.97\% & 98.66\%$\pm$0.42\% & $<$0.01\% & 6.82\%$\pm$2.45\% & 99.98\%$\pm$0.02\% & 53.23\%$\pm$14.94\% \\
    \cifarten{} & 95.21\%$\pm$0.13\% & 98.57\%$\pm$0.66\% & 92.93\%$\pm$0.39\% & $<$0.01\%  & $<$0.01\% & 99.98\%$\pm$0.02\% & 99.99\%$\pm$0.01\% \\
    \gtsrb{}    & 99.38\%$\pm$0.19\% & 20.17\%$\pm$1.48\% & 96.99\%$\pm$1.45\% & 1.20\%$\pm$0.27\% & 1.35\%$\pm$0.27\% & 52.86\%$\pm$9.05\% & 48.26\%$\pm$4.68\% \\
  \bottomrule
  \end{tabular}
  \caption{\label{tab:nmldnns}The performance of topologically manipulated
    \dnn{}s compared to standard \dnn{}s. For standard \dnn{}s, we report the
    average and standard deviation of the (benign) accuracy and the targeting
    success rate (\tsr{}). \tsr{} is defined as the rate at which the \dnn{}
    emitted the target class on a transferred adversarial example. For
    topologically manipulated \dnn{}s, we report the average and standard
    deviation of the accuracy, the \tsr{}, and the manipulation success rate
    (\msr{}). \msr{} is the rate at which adversarial
    examples were classified as specified by the derangements drawn at training
    time. \tsr{} and \msr{} are reported for adversarial examples produced 
    against the same \dnn{}s used during training or ones produced 
    against held-out (\emph{h/o}) \dnn{}s.\lujo{it would be nice to align the columns on decimal points}} 
\end{table*}

To measure the benign accuracy, we classified the original (benign) samples from
the test sets of datasets using the 18 topologically manipulated \dnn{}s
as well as the (better-performing) standard \dnn{}s that we held out from training topologically
manipulated \dnn{}s (i.e., architectures 1--3). \tabref{tab:nmldnns} reports the
average and standard deviation of the benign accuracy. Compared to the standard
\dnn{}s, the topologically manipulated ones had only slightly lower accuracy
(0.64\%--2.39\% average decrease in accuracy). Hence, we can conclude that
topologically manipulated \dnn{}s were accurate. \lujo{why compare
  against different architectures and not the same ones here? this is
  not yet about attackers; it's about performance in a benign setting}
\mahmood{because we want to be able to evaluate the TR for transferred attacks
  in the table, see next couple of paragraphs. we're setting a higher bar to
  ourselves by doing the comparison this way.}

Next, we measured the extent to which topology manipulation was successful. To
this end, we computed adversarial examples for the \dnn{}s used to
train topologically manipulated \dnn{}s (i.e., architectures 4--6) or\lujo{when
  is it which one?}\mahmood{added a reminder} \dnn{}s held out from training
(i.e., architectures 1--3). Again, we used \pgd{} and techniques to enhance
transferability to
compute the adversarial examples. As in prior work, we set $\epsilon{}=0.3$ for \mnist{} and
$\epsilon{}=\frac{8}{255}$ for \cifarten{} and \gtsrb{}, and we ran \pgd{} for 40
iterations. For each benign sample,
$x$, we created $\nclasses{}-1$ corresponding adversarial examples, one targeting
every class $c_t\neq{}c_x$. To reduce the computational load, we used a random
subset of benign samples from the test sets: 1024 samples for \mnist{} and 512
samples for the other datasets.

For constructing robust \nML{} ensembles, the topologically manipulated \dnn{}s
should classify adversarial examples as specified during training, or, at least,
differently than the adversary anticipates. We estimated the former via the
\emph{manipulation success rate} (\msr{})---the rate at which adversarial examples were
classified as specified by the derangements drawn at training time---while we
estimated the latter via the \emph{targeting success rate} (\tsr{})---the rate at which
adversarial examples succeeded at being misclassified as the target class. A
successfully trained topologically manipulated \dnn{} should obtain a high
\msr{} and a low \tsr{}.

\tabref{tab:nmldnns} presents the average and standard deviation of \tsr{}s and
\msr{}s for topologically manipulated \dnn{}s, as well as the \tsr{}s for
standard \dnn{}s. One can immediately see that targeting was much less likely to
succeed for a topologically manipulated \dnn{} (average \tsr{}$\le$6.82\%) than
for a standard \dnn{} (average \tsr{}$\ge$20.17\%, and as high as 98.57\%). In 
fact, across all datasets and regardless of whether the adversarial examples were computed
against held-out \dnn{}s, the likelihood of targeting to succeed
for standard \dnn{}s was $\times{}$6.31 or higher than for topologically manipulated \dnn{}s. This
confirms that the adversarial directions of topologically manipulated \dnn{}s
were vastly different than those of standard \dnn{}s. Furthermore, as reflected
in the \msr{} results, we found that topologically manipulated \dnn{}s were
likely to classify adversarial examples as specified by their corresponding
derangements. Across all datasets, and regardless of whether the adversarial examples
were computed against held-out \dnn{}s or not, the average likelihood of topologically
manipulated \dnn{}s to classify adversarial examples according to specification
was $\ge$48.26\%.\lujo{that average doesn't seem to depend on the
  dataset or held-out dnns, right?}\mahmood{good catch. fixed.}
For example, an average of 99.99\% of the adversarial examples
produced against the held-out \dnn{}s of \cifarten{} were classified according to
specification by the topologically manipulated \dnn{}s.

In summary, the results indicate that the topologically manipulated \dnn{}s
satisfied their objectives to a large extent: they accurately classified benign
inputs, and their topology with respect to adversarial directions was different
than that of standard \dnn{}s, as they often classified adversarial examples
according to the specification that was selected at training time.

\subsection{Evaluating \nML{} Ensembles}
\label{sec:res:nml}
\lujo{for clarity, maybe split this up into a subsec on experiment
  setup and then individual subsecs on black/gray/white-box
  evaluations. l2 and overhead would probably need to be broken out
  the same way then, which isn't ideal. maybe changing parheadings to
  black would be similarly effective at providing structure at a
  glance}
\mike{I tried adding in some subsections and parheadings, though it
  can probably be improved}

Now we describe our evaluation of \nML{} ensembles. We begin by describing the
experiment setup and reporting the benign accuracy of ensembles of various
sizes and thresholds. We then present experiments to evaluate \nML{} ensembles
in various settings and compare \nML{} against other defenses. We finish with
an evaluation of the time overhead incurred when deploying \nML{} for inference.

\subsubsection{Setup}
\mike{In a perfect world, we'd change this to be past tense.  Hesitant
to suggest we make such a big change right now, though.}
The \nML{} ensembles that we constructed were composed of the topologically
manipulated \dnn{}s described in the previous section. Particularly, we constructed
ensembles containing five (\nML[5]), nine (\nML[9]), or 18 (\nML[18])
\dnn{}s, as we found ensembles of certain sizes to be more suitable than others
at balancing benign accuracy, security, and inference time
in different settings. For number of variants $\nvariants{} \le 9$, we
selected \dnn{}s whose derangements disagreed in all indices for the ensembles. For
$\nvariants=18$, we selected all the \dnn{}s. Note that since \gtsrb{} contains a
large number of classes ($\nclasses=43$),  we could train 18 \dnn{}s with
derangements that disagreed on all indices. However, we avoided doing so to save
compute cycles, as the \dnn{}s that we trained performed well despite having
derangements that agreed on certain indices.

\parheading{Other Defenses} We compared \nML{} with three state-of-the-art defenses:
\advpgd{}~\cite{Madry17AdvTraining, Shafahi19PGD},
\lid{}~\cite{Ma18LID}, and \nic{}~\cite{Ma19Detection}.

\advpgd{} and \lid{} use adversarial examples at
training time; we set the magnitude of the \lpnorm{\infty}-norm
perturbations to $\epsilon{}=0.3$ to produce adversarial examples for
\mnist{}, and to $\epsilon{}=\frac{8}{255}$ for \cifarten{} and
\gtsrb{}, as these are typical attack magnitudes that defenses attempt
to prevent (e.g.,~\cite{Madry17AdvTraining, Papernot18DkNN}).

For \advpgd{}, we 
implemented and used the \emph{free adversarial training} method of Shafahi et
al.~\cite{Shafahi19PGD}, which adversarially trains \dnn{}s in the
same amount of time as standard training.  We used \advpgd{} to train four
defended \dnn{}s for each dataset---one to be used by the defender, and three to 
be used for transferring attacks in the grey-box setting (see below). To give 
the defense an advantage, we used the best performing architecture for the
defender's \dnn{} (architecture 1 from \tabref{tab:archs}), and the least
performing architectures for the attacker's \dnn{}s (architectures 4--6). For
\cifarten{}, as the \dnn{} that we obtained after training did not perform as
well as prior works', we used the adversarially trained \dnn{} released by
Madry et al.~\cite{Madry17AdvTraining} as the defender's \dnn{}.

For training \lid{} detectors, we used the implementation that was published by
the authors~\cite{Ma18LID}. As described in \secref{sec:relwork}, \lid{} detectors
compute \lid{} statistics for intermediate representations of inputs and feed
the statistics to a logistic regression classifier to detect adversarial
examples. The logistic regression classifier is trained using \lid{} statistics
of benign samples, adversarial examples, and noisy variants of benign samples
(created by adding non-adversarial Gaussian noise). We tuned the amount of 
noise for best performance (i.e., highest benign accuracy and detection rate of
adversarial examples). For \cifarten{} and \gtsrb{}, we trained \lid{} detectors
for \dnn{}s of architecture 1. For \mnist{}, we trained a \lid{} detector for
the \dnn{} architecture that was used in the original work---architecture 4.

Using code that we obtained directly from the authors, we trained two \nic{}
detectors per dataset---one to be used by the defender (in all settings), and one by the attacker
in the grey-box setting. The defender's detectors were trained for \dnn{}s of the
same architectures as for \lid{}. For the attacker, we trained detectors for
\dnn{}s of architectures 1, 4, and 2, for \mnist{}, \cifarten{}, and \gtsrb{},
respectively. We selected the attackers' \dnn{} architectures arbitrarily, and
expect that other architectures would perform roughly the same. The \ocsvm{}s
that we trained for \nic{} have Radial Basis Functions (RBF) as kernels, since
these were found to perform best for detection~\cite{Ma19Detection}.

\parheading{Attack Methodology}
We evaluated \nML{} and other defenses against untargeted attacks, as they are
easier to attain from the point of view of attackers, and more challenging to
defend against from the point of view of the defender. For \advpgd{}, \lid{},
and \nic{}, we used typical \pgd{} untargeted attacks with various adaptations
depending on the setting. Typical
untargeted attacks, however, are unlikely to evade \nML{} ensembles, as if the target is
not specified by the attack then each \dnn{} in the ensemble may classify the
resulting adversarial example differently, thus detecting the presence of an
attack. To address this, we used a more powerful attack, similarly to Carlini and
Wagner~\cite{Carlini17Robustness}. The attack builds on targeted
\pgd{} to generate adversarial examples targeting all possible incorrect
classes (i.e., $\nclasses{}-1$ in total) and checks if one of these adversarial
examples is misclassified by a large number of \dnn{}s in the ensemble
($\ge\tau{}$), and so is not detected as adversarial by the \nML{}
ensemble. Because targeting every possible class increases the computational
load of attacks, we used random subsets of test sets to produce adversarial
examples against \nML{}. In particular, we used 1,000 samples for \mnist{} and
512 samples for \cifarten{} and \gtsrb{}. The magnitude of \lpnorm{\infty}-norm
of perturbations that we considered were $\epsilon{}\le{}0.3$ for \mnist{} and
$\epsilon{}\le{}\frac{8}{255}$ for \cifarten{} and \gtsrb{}, and we ran attacks
for 40 iterations. Again, we used techniques to attack ensembles and enhance the
transferability of attacks~\cite{Dong19TI, Liu17Transfer, Tramer18Ensemble, Xie19ID}.

When directly attacking a \dnn{} defended by \lid{}, we simply produced
adversarial examples that were misclassified with high confidence against the
\dnn{} while ignoring the defense. This approach is motivated by prior work,
which found that high confidence adversarial examples mislead \lid{} with high
likelihood~\cite{Athalye18Attack}. When attacking a \dnn{} defended by
\nic{}, we created a new \dnn{} by combining the logits of the original \dnn{}
and those of the classifiers built on top of every intermediate layer. We found
that forcing the original \dnn{} and the intermediate classifiers to (mis)classify
adversarial examples in the same manner often led the \ocsvm{}s to misclassify
adversarial examples as benign.

\parheading{Measures}
In the context of adversarial examples, a defense's success is measured in its
ability to prevent adversarial examples, while maintaining high benign accuracy
(e.g., close to that of a standard classifier). The benign accuracy is the
rate at which benign inputs are classified correctly and not detected as
adversarial. In contrast, the success rate of attacks is indicative of the
defense's ability to prevent adversarial examples (high success rate indicates a
weak defense, and vice versa). For untargeted attacks, the success rate can be
measured by the rate at which adversarial examples are not detected and are
classified to a class other than the true class. Note that \advpgd{} is a method
for robust classification, as opposed to detection, and so adversarially trained
\dnn{}s always output an estimation of the most likely class (i.e., abstaining
from classifying an input that is suspected to be adversarial is not an option).

We tuned the defenses at inference time to compute different tradeoffs on the
above metrics. In the case of \nML{}, we computed the benign accuracy and
attacks' success rates for threshold values
$\lceil{}\frac{\nvariants{}+1}{2}\rceil\le{}\tau{}\le{}\nvariants{}$. For
\lid{}, we evaluated the metrics for different thresholds $\in[0,1]$ on the 
logistic regression's probability estimates that inputs are adversarial. \nic{}
emits scores in arbitrary ranges---the higher the score the more likely the
input to be adversarial. We computed accuracy and success rate tradeoffs of
\nic{} for thresholds between the minimum and maximum values emitted for benign
samples and adversarial examples combined. In all cases, both the benign
accuracy and attacks' success rates decreased as we increased the thresholds.
\advpgd{} results in a single model that cannot be tuned at inference time. We
report the single operating point that it achieves.

\subsubsection{Results}
We now present the results of our evaluations, in terms of benign
accuracy; resistance to adversarial examples in the black-,
grey-, and white-box settings; and overhead to classification
performance.

\parheading{Benign Accuracy}
We first report on the benign accuracy of \nML{} ensembles. In
particular we were interested in finding how different was the accuracy of
ensembles from single standard \dnn{}s. Ideally, it is desirable to maintain
accuracy that is as close to that of standard training as possible.

\begin{figure*}[h!]
  \centering
  \begin{subfigure}{0.3\textwidth}
    \centering
    \includegraphics[width=\textwidth]{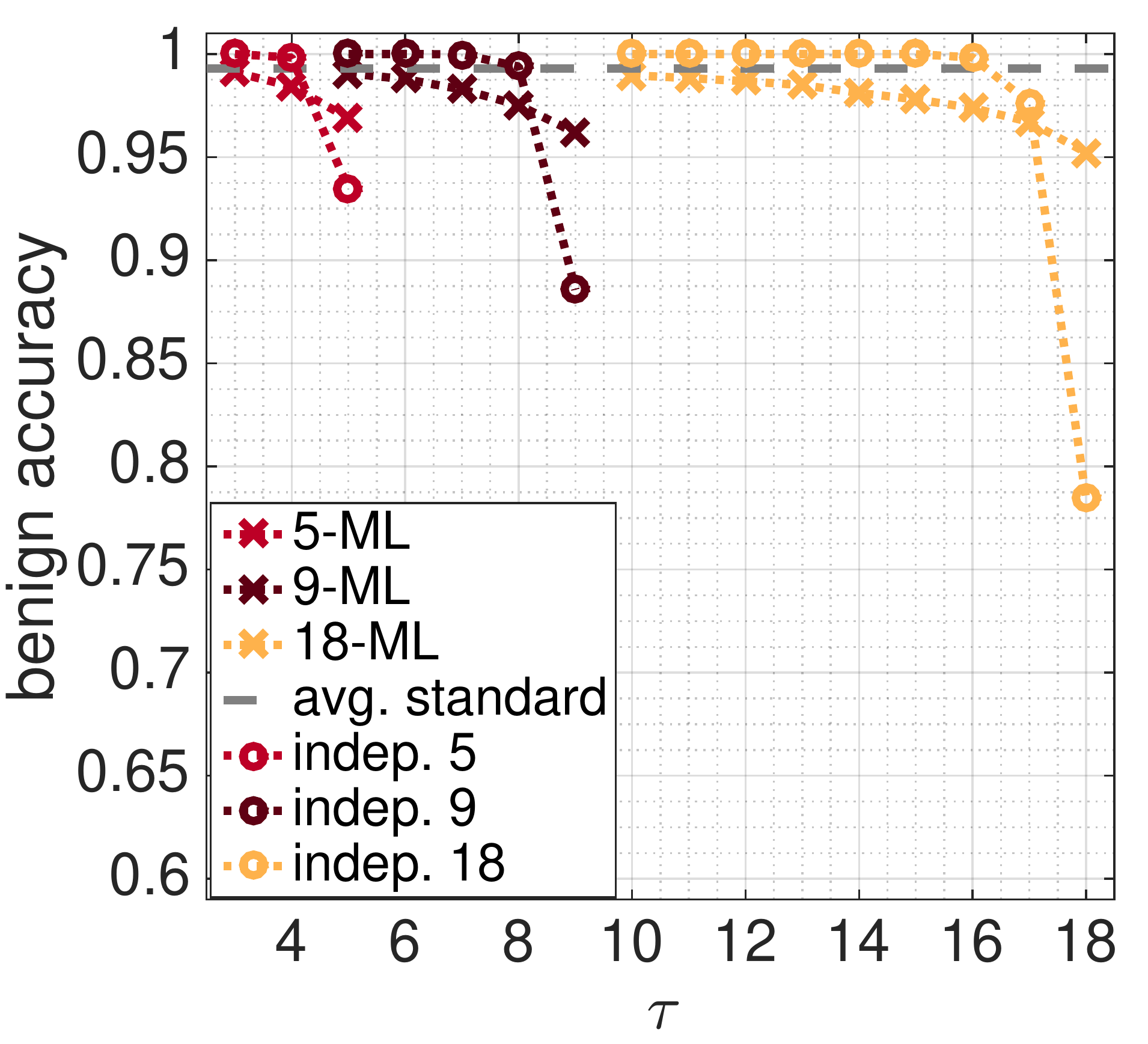}
    \caption{\mnist{}}
  \end{subfigure}
  \begin{subfigure}{0.3\textwidth}
    \centering
    \includegraphics[width=\textwidth]{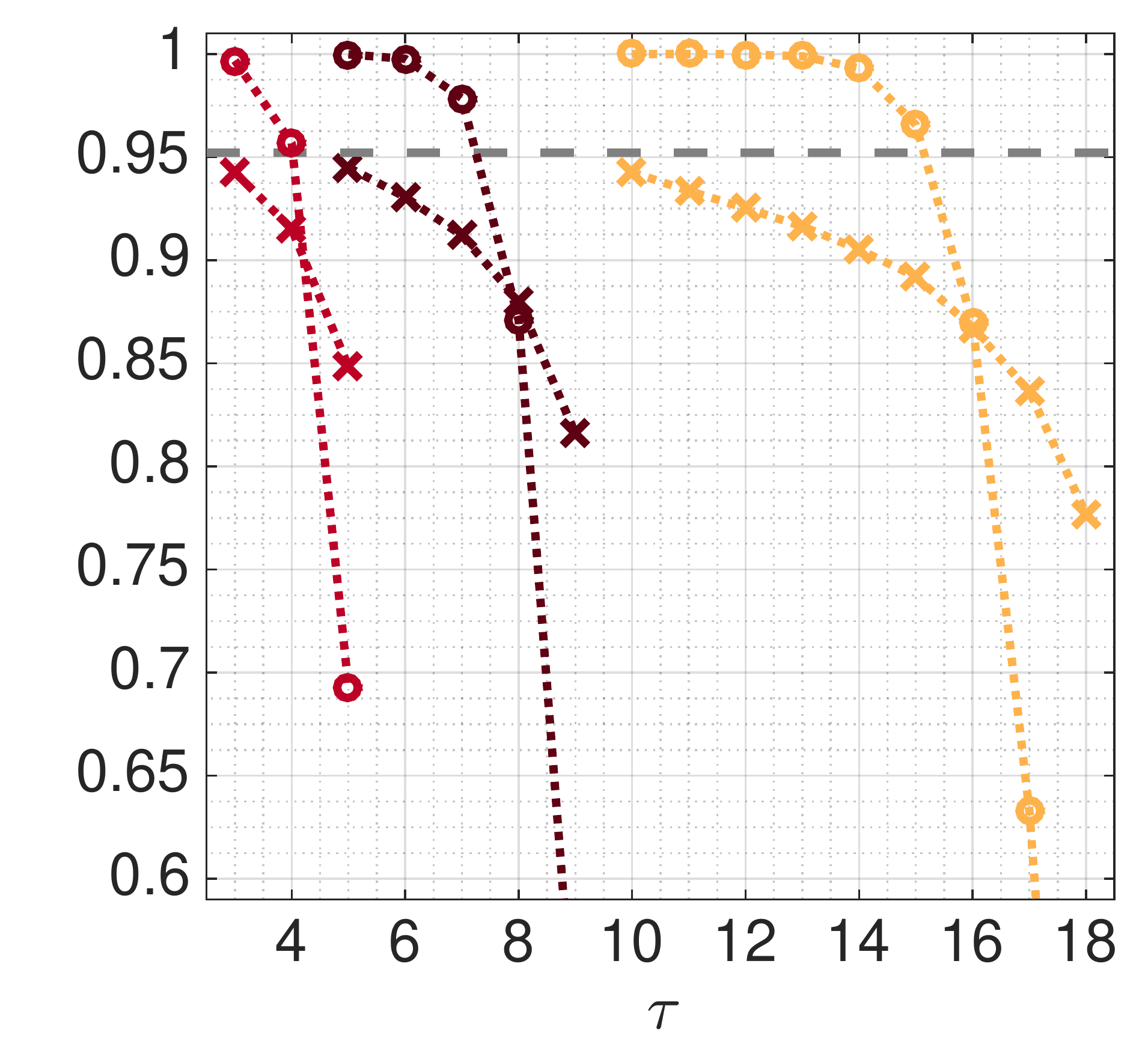}
    \caption{\cifarten{}}
  \end{subfigure}
  \begin{subfigure}{0.3\textwidth}
    \centering
    \includegraphics[width=\textwidth]{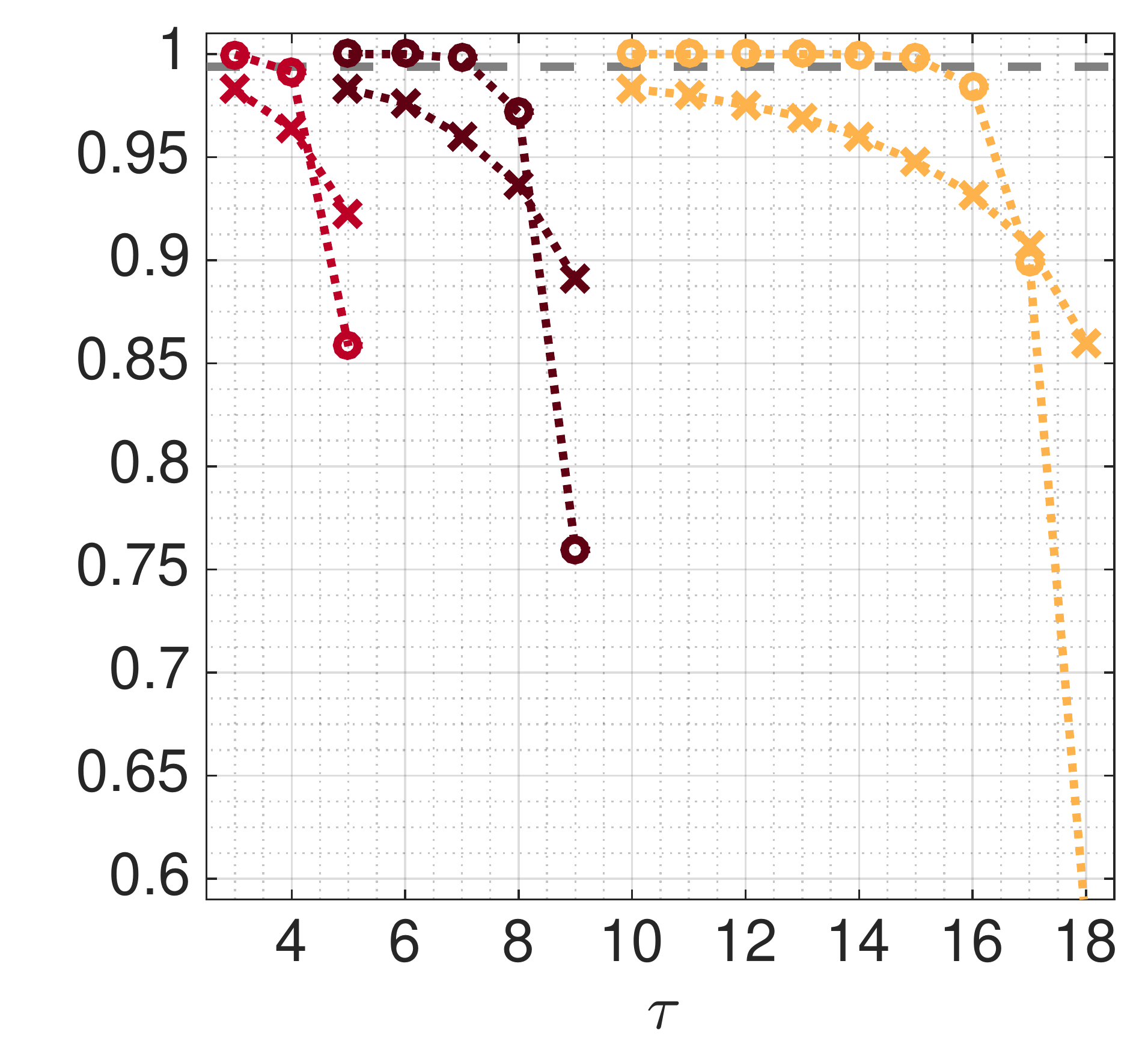}
    \caption{\gtsrb{}}
  \end{subfigure}
  \caption{\label{fig:accthresh}The benign accuracy of \nML{} ensembles of different
    sizes as we varied the thresholds. For reference, we show the average accuracy
    of a single standard \dnn{} (avg.\ standard), as well as the accuracy of
    hypothetical ensembles whose constituent \dnn{}s are assumed to have
    independent errors and the average accuracy of topologically manipulated
    \dnn{}s (indep.\ $\nvariants$). The dotted lines connecting the markers were added
    to help visualize trends, but do not correspond to actual operating points.
    \lujo{in the first row figures, I think nML would stand out more if all the nML lines were red, say with an x and an o and a box for points. We could reuse the same line color and shape for nML in the second row of graphs for consistency.}\mahmood{done}
  }
\end{figure*}

\figref{fig:accthresh} compares \nML{} ensembles' accuracy with standard \dnn{}s,
as well as with hypothetical ensembles whose \dnn{} members have the average
accuracy of topologically manipulated \dnn{}s and independent errors. For low
thresholds, it can be seen that the accuracy of \nML{} was close to the average
accuracy of standard benign \dnn{}s. As we increased the thresholds, the accuracy
decreased. Nonetheless, it did not decrease as dramatically as for ensembles
composed from \dnn{}s with independent error. For example, the
accuracy of an ensemble containing five independent \dnn{}s each with an
accuracy of 92.93\% (the average accuracy of topologically manipulated \dnn{}s
on \cifarten{}) is 69.31\% when $\tau{}=5$ (i.e., we require all \dnn{}s to
agree). In comparison, \nML[5] achieved 84.82\% benign accuracy for the same
threshold on \cifarten{}.

We can conclude that \nML{} ensembles were almost as accurate as standard models
for low thresholds, and that they did not suffer from dramatic accuracy loss as
thresholds were increased.

\parheading{Black-box Attacks}
In the black-box setting, as the attacker is unaware of the use of defenses and has
no access to the classifiers, we used non-interactive transferability-based
attacks to transfer adversarial examples produced against standard surrogate
models. For \nML{} and \advpgd{}, we used a strong attack by transferring
adversarial examples produced against the standard \dnn{}s held-out from
training the defenses. For \lid{} and \nic{} we found that transferring
adversarial examples produced against the least accurate standard \dnn{}s
(architecture 6) was sufficient to evade classification with high success rates.

\begin{figure*}[h!]
  \centering
  \begin{subfigure}{0.3\textwidth}
    \centering
    \includegraphics[width=\textwidth]{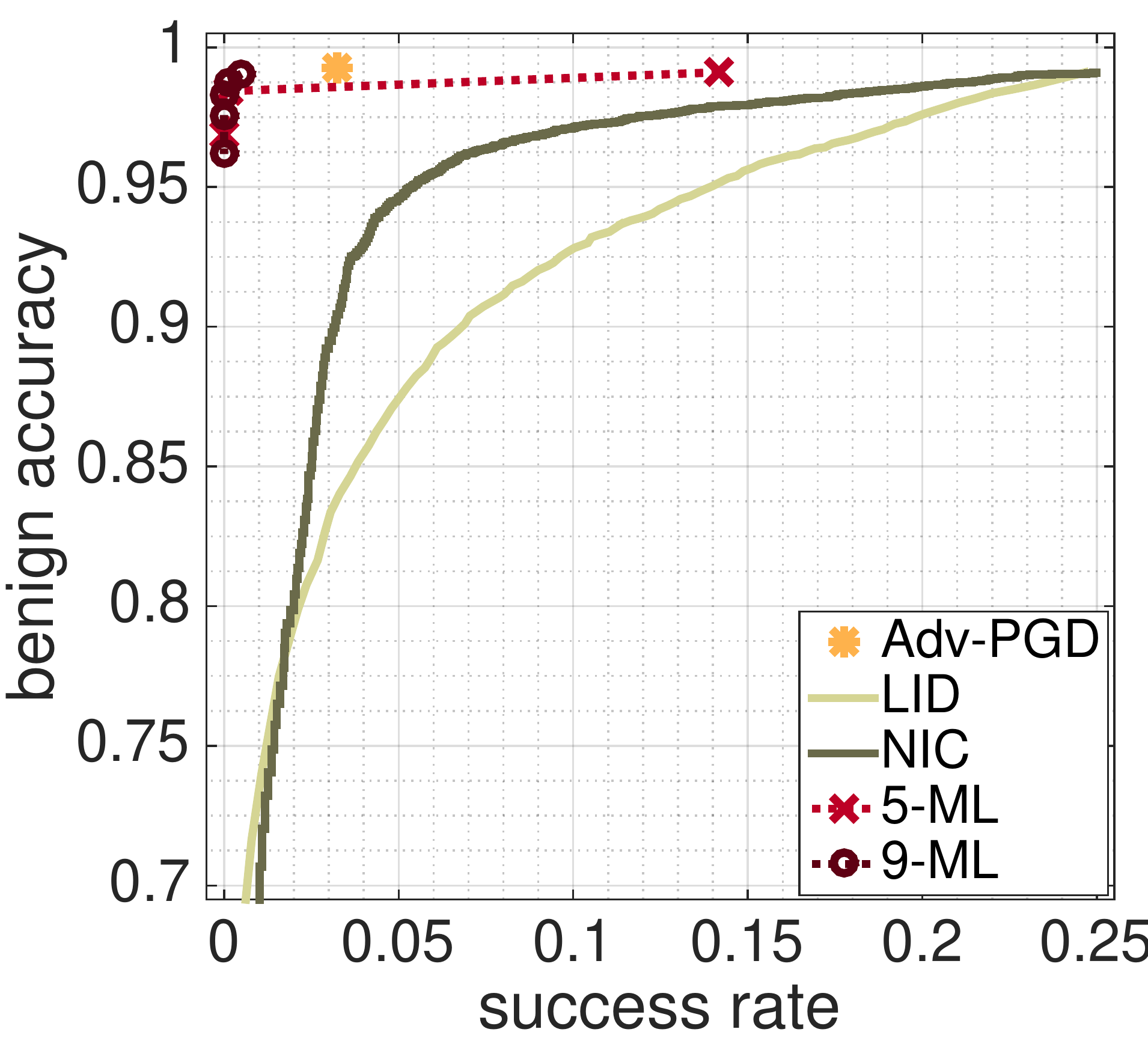}
    \caption{\mnist{}}
  \end{subfigure}
  \begin{subfigure}{0.3\textwidth}
    \centering
    \includegraphics[width=\textwidth]{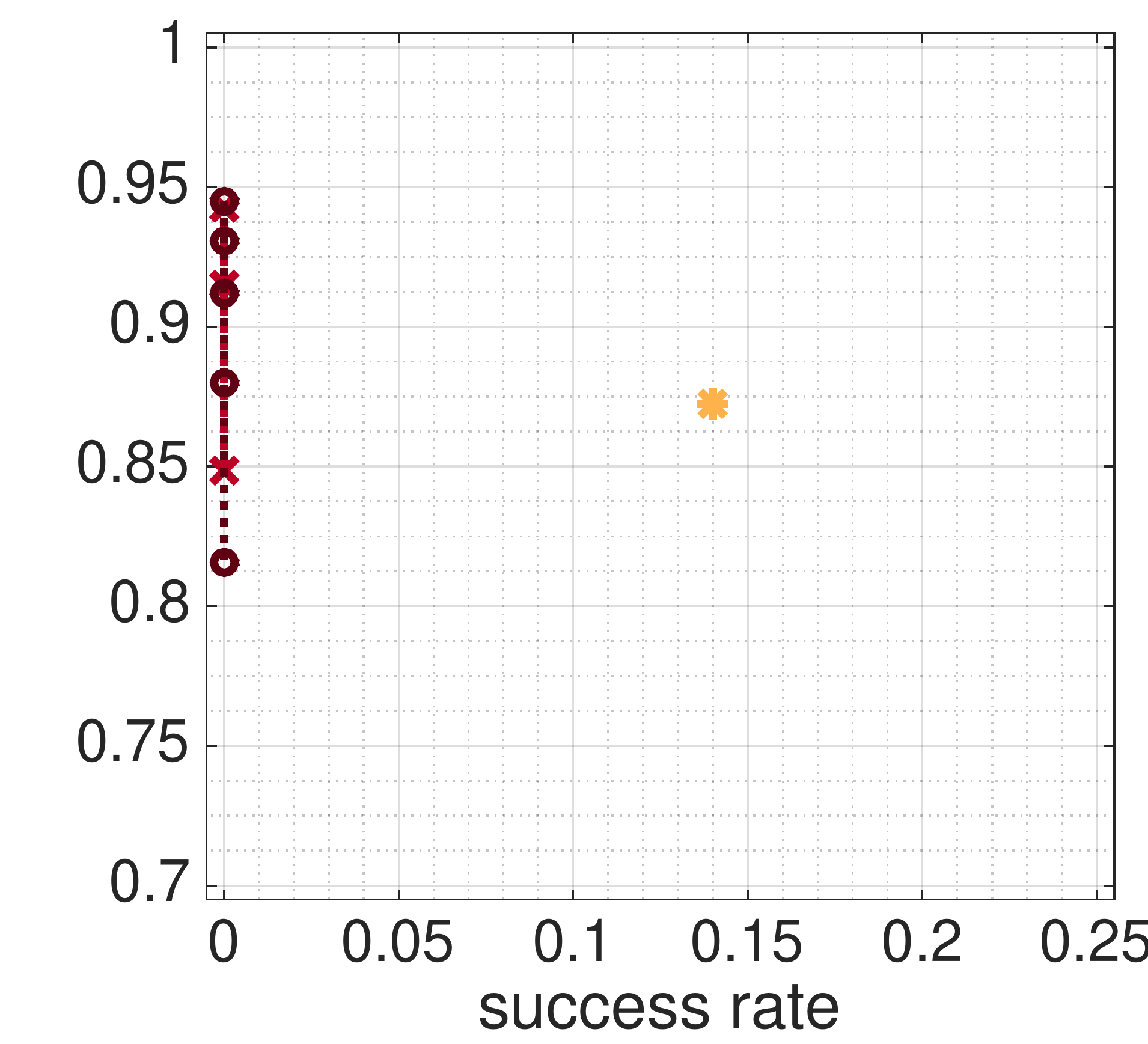}
    \caption{\cifarten{}}
  \end{subfigure}
  \begin{subfigure}{0.3\textwidth}
    \centering
    \includegraphics[width=\textwidth]{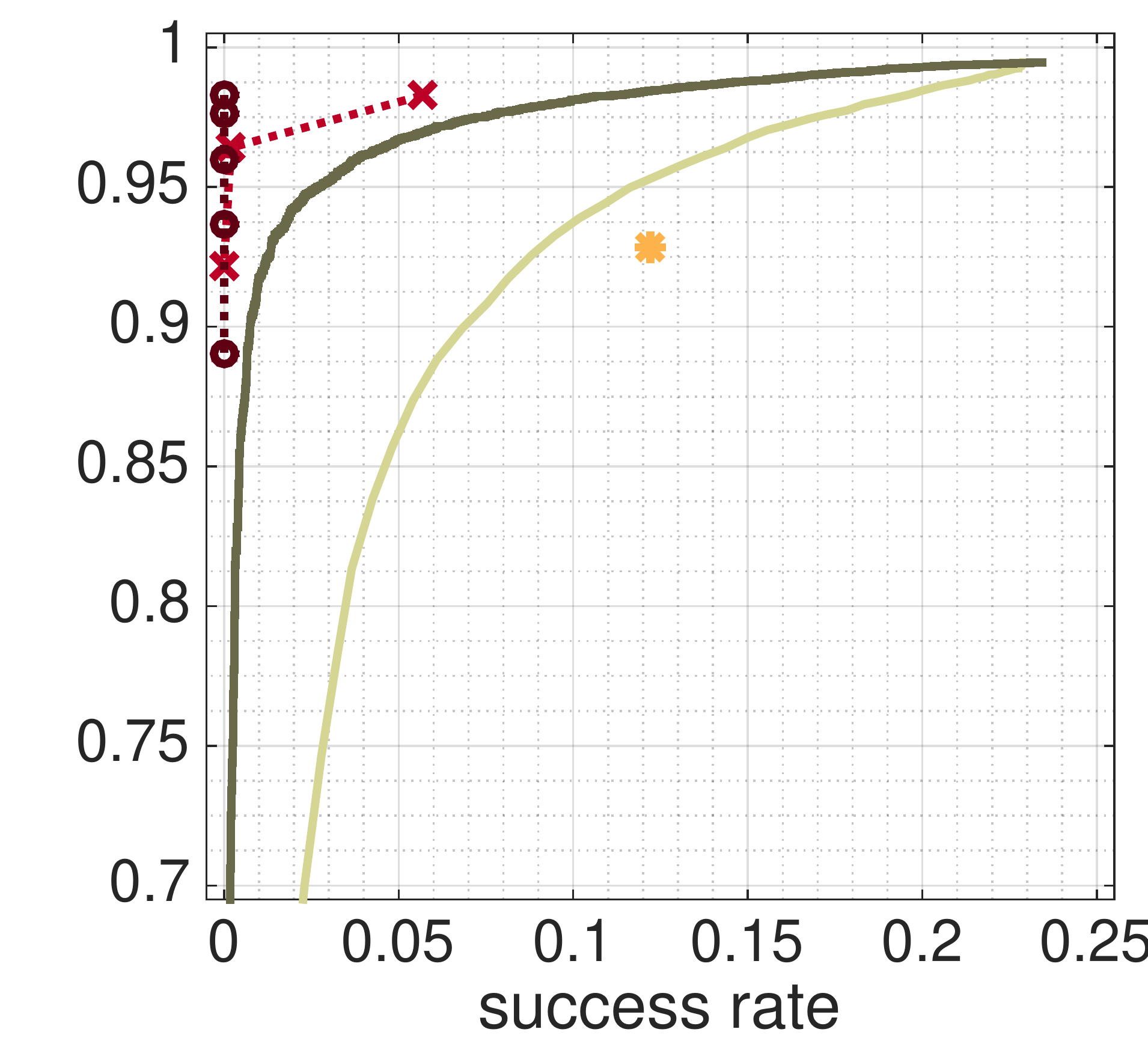}
    \caption{\gtsrb{}}
  \end{subfigure}
  \caption{\label{fig:bboxcomp}Comparison of defenses' performance in the
    \emph{black-box setting}. The \lpnorm{\infty}-norm of perturbations was set to
    $\epsilon{}=0.3$ for \mnist{} and $\epsilon{}=\frac{8}{255}$ for \cifarten{}
    and \gtsrb{}. Due to poor performance, the \lid{} and \nic{} curves were
    left out from the \cifarten{} plot after zooming in. The dotted lines connecting
    the \nML{} markers were added to help visualize trends, but do not correspond to
    actual operating points.}
\end{figure*}

\figref{fig:bboxcomp} summarizes the results for the attacks with the highest
magnitudes. For \nML{}, we report the performance of \nML[5] and \nML[9], which
we found to perform well in the black-box setting (i.e., they achieved high
accuracy while mitigating attacks). It can be seen that \nML{}
outperformed other defenses across all the datasets. For example, for
\cifarten{}, \nML[9] was able to achieve 94.50\% benign accuracy at 0.00\% attack
success rate. In contrast, the second best defense, \advpgd{}, achieved 87.24\%
accuracy at 14.02\% attack success rate.

While not shown for the sake of brevity, additional experiments that we performed
demonstrated that \nML{} in the black-box setting performed roughly the same as in
\figref{fig:bboxcomp} when \emph{1)} different perturbation magnitudes were used
($\epsilon\in\{0.05, 0.1, \dots, 0.4\}$ for \mnist{} and
$\epsilon\in\{\frac{1}{255}, \frac{2}{255}, \dots, \frac{8}{255}\}$
for \cifarten{} and \gtsrb{}); \emph{2)} individual standard
\dnn{}s were used as surrogates to produce adversarial examples; and \emph{3)}
the same \dnn{}s used to train topologically manipulated \dnn{}s were used as
surrogates.

\parheading{Grey-box Attacks}
In the grey-box setting (where attackers are assumed to be aware of the
deployment of defenses, but have no visibility to the parameters of
the classifier and the defense), we attempted to transfer adversarial
examples produced against surrogate defended classifiers. For \nML{},
we evaluated \nML[5] and 
\nML[9] in the grey-box setting. As surrogates, we used \nML{}
ensembles of the same sizes and architectures to produce adversarial
examples. The derangements used for training the \dnn{}s of the surrogate
ensembles were picked independently from the defender's ensembles
(i.e., the derangements could agree with the defender's derangements on
certain indices). For \advpgd{}, we used three adversarially trained
\dnn{}s different than the defender's \dnn{}s as surrogates. For \nic{} we used
standard \dnn{}s and corresponding detectors (different than the
defender's, see above for  training details) as surrogates. For
\lid{}, we simply produced adversarial examples that were
misclassified with high confidence against undefended standard \dnn{}s of 
architecture 2 (these were more architecturally similar to the defenders' \dnn{}s
than the surrogates used in the black-box setting).

\begin{figure*}[h!]
  \centering
  \begin{subfigure}{0.3\textwidth}
    \centering
    \includegraphics[width=\textwidth]{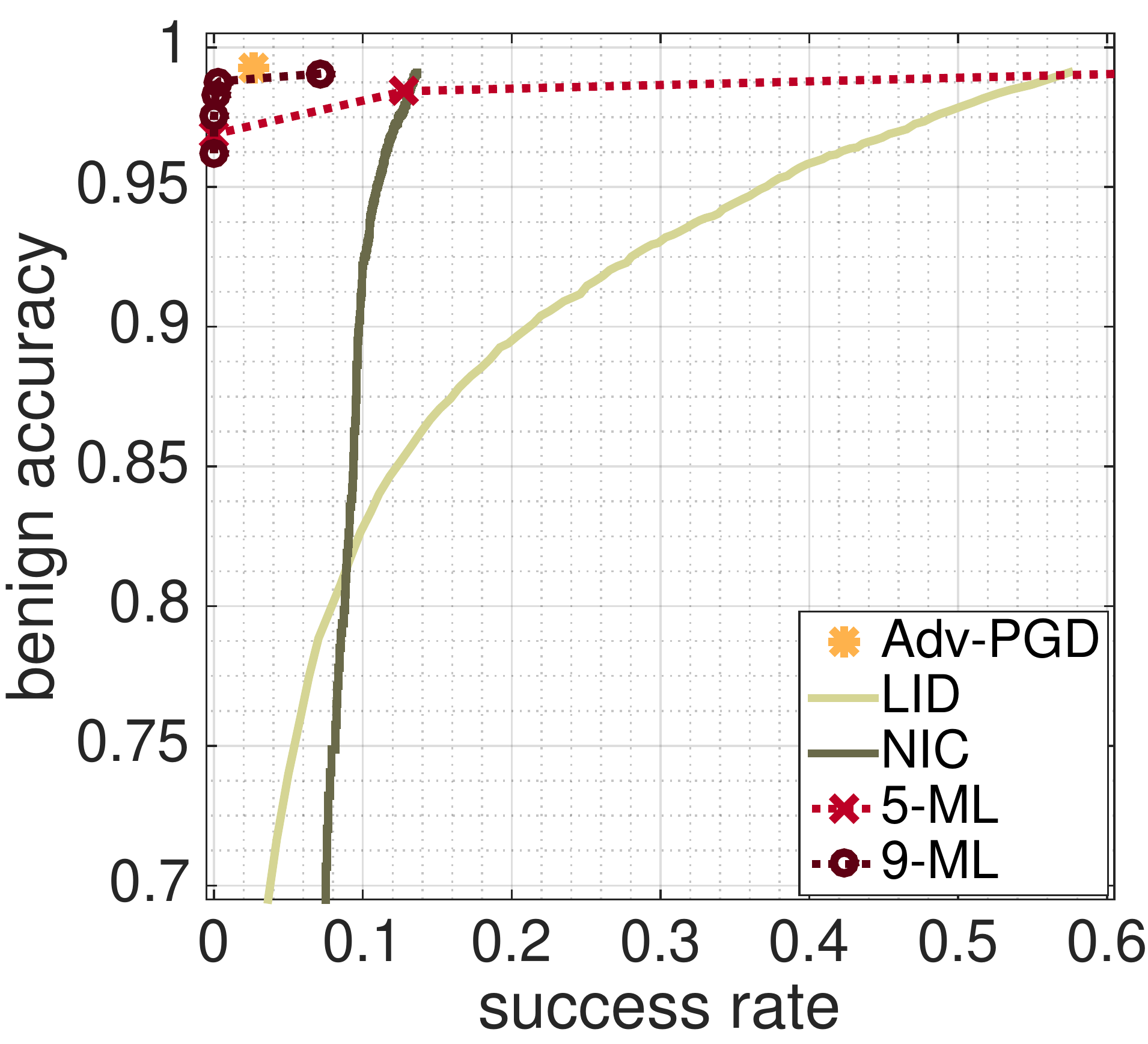}
    \caption{\mnist{}}
  \end{subfigure}
  \begin{subfigure}{0.3\textwidth}
    \centering
    \includegraphics[width=\textwidth]{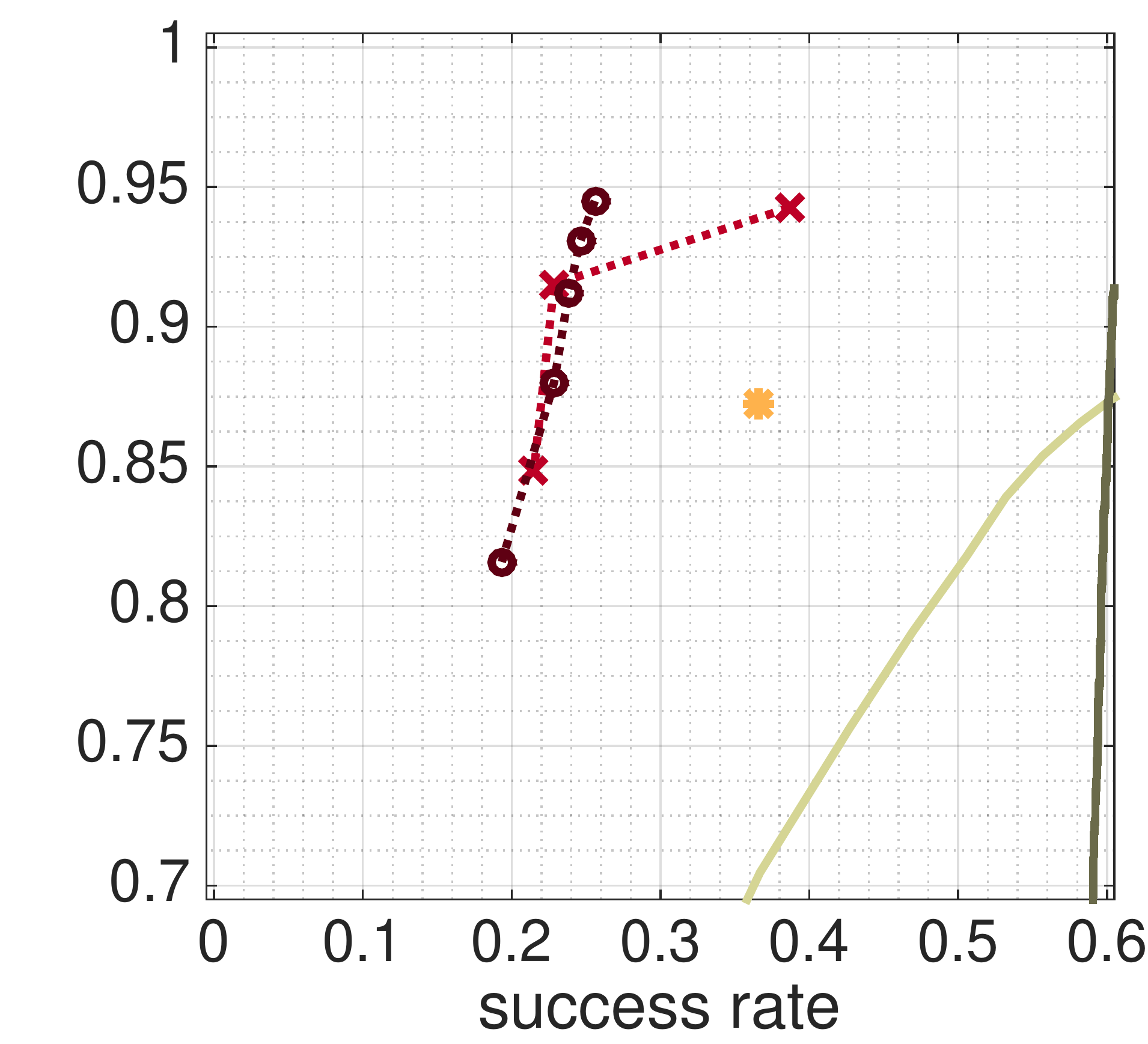}
    \caption{\cifarten{}}
  \end{subfigure}
  \begin{subfigure}{0.3\textwidth}
    \centering
    \includegraphics[width=\textwidth]{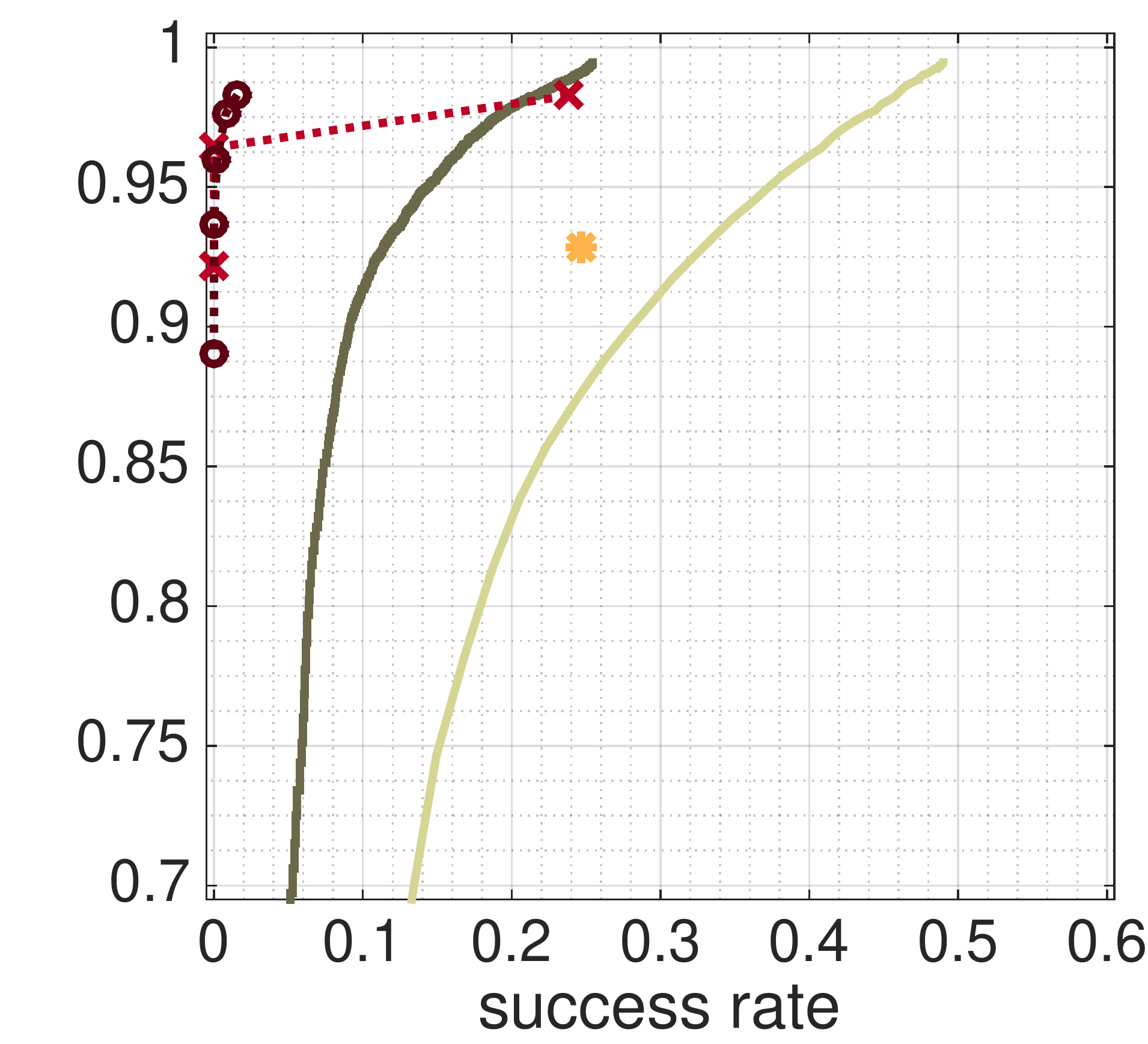}
    \caption{\gtsrb{}}
  \end{subfigure}
  \caption{\label{fig:gboxcomp}Comparison of defenses' performance in the
    \emph{grey-box setting}. The \lpnorm{\infty}-norm of perturbations was set to
    $\epsilon{}=0.3$ for \mnist{} and $\epsilon{}=\frac{8}{255}$ for \cifarten{}
    and \gtsrb{}. The dotted lines connecting the \nML{} markers were added to
    help visualize trends, but do not correspond to actual operating points.}
\end{figure*}

\figref{fig:gboxcomp} presents the performance of the defenses against the
attacks with the highest magnitudes. Again, we found \nML{} to achieve favorable
performance over other defenses. In the case of \gtsrb{}, for instance, \nML[9]
could achieve 98.30\% benign accuracy at 1.56\% attack success rate. None of the
other defenses was able to achieve a similar accuracy while preventing $\ge$98.44\%
of the attacks for \gtsrb{}.

Additional experiments that we performed showed that \nML{} maintained
roughly the same performance as we varied the number of \dnn{}s in the
attacker's surrogate ensembles ($\nvariants{}\in\{1, 5, 9\}$) and the
attacks' magnitudes 
($\epsilon\in\{0.05, 0.1, \dots, 0.4\}$ for \mnist{} and
$\epsilon\in\{\frac{1}{255}, \frac{3}{255}, \frac{5}{255}, \frac{7}{255}\}$
for \cifarten{} and \gtsrb{}).

\parheading{White-Box Attacks}
Now we turn our attention to the white-box setting, where attackers are assumed
to have complete access to classifiers' and defenses' parameters. In this 
setting, we leveraged the attacker's knowledge of the classifiers' and defenses'
parameters to directly optimize the adversarial examples against them.

\begin{figure*}[h!]
  \centering
  \begin{subfigure}{0.3\textwidth}
    \centering
    \includegraphics[width=\textwidth]{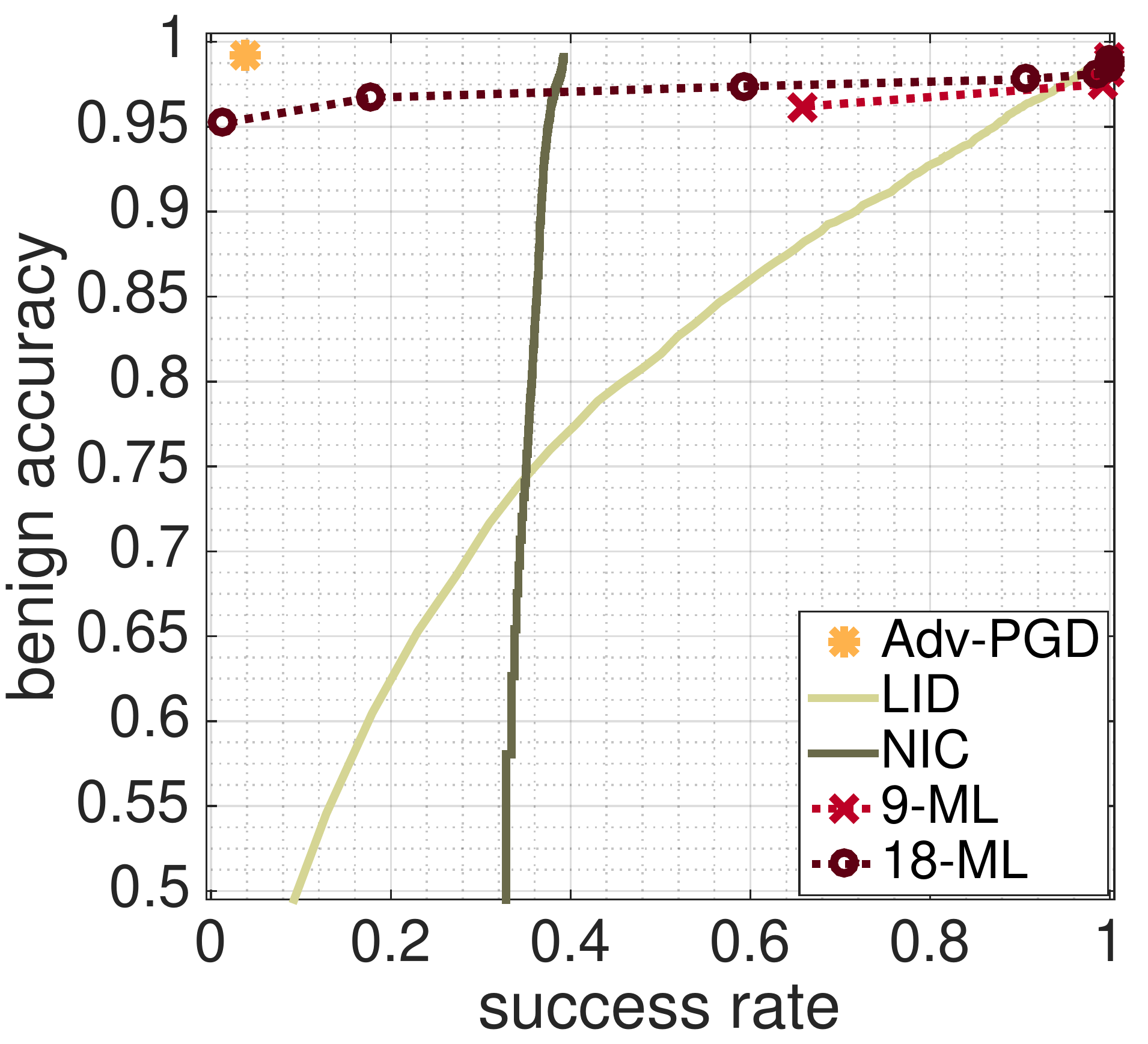}
    \caption{\mnist{}}
  \end{subfigure}
  \begin{subfigure}{0.3\textwidth}
    \centering
    \includegraphics[width=\textwidth]{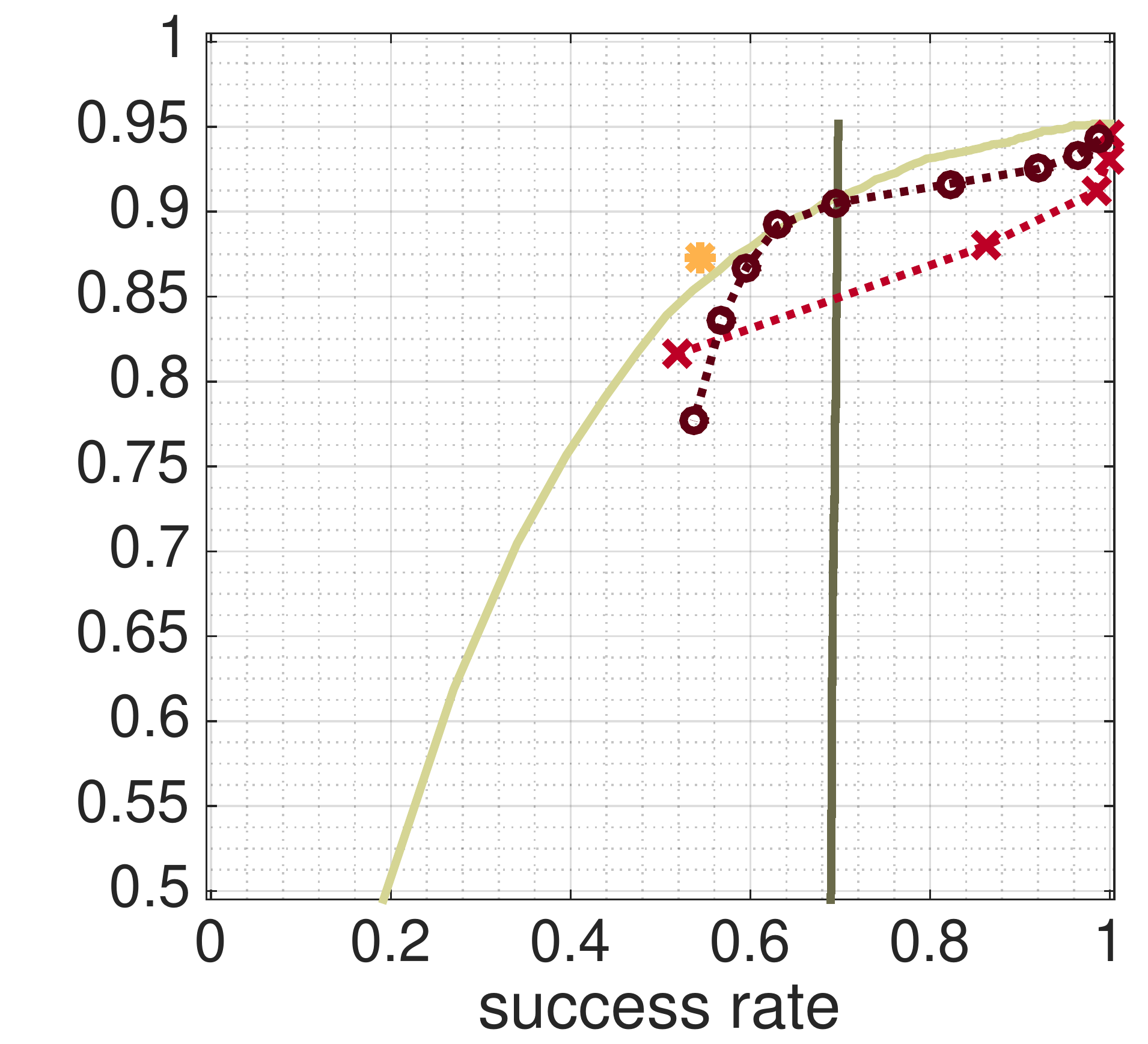}
    \caption{\cifarten{}}
  \end{subfigure}
  \begin{subfigure}{0.3\textwidth}
    \centering
    \includegraphics[width=\textwidth]{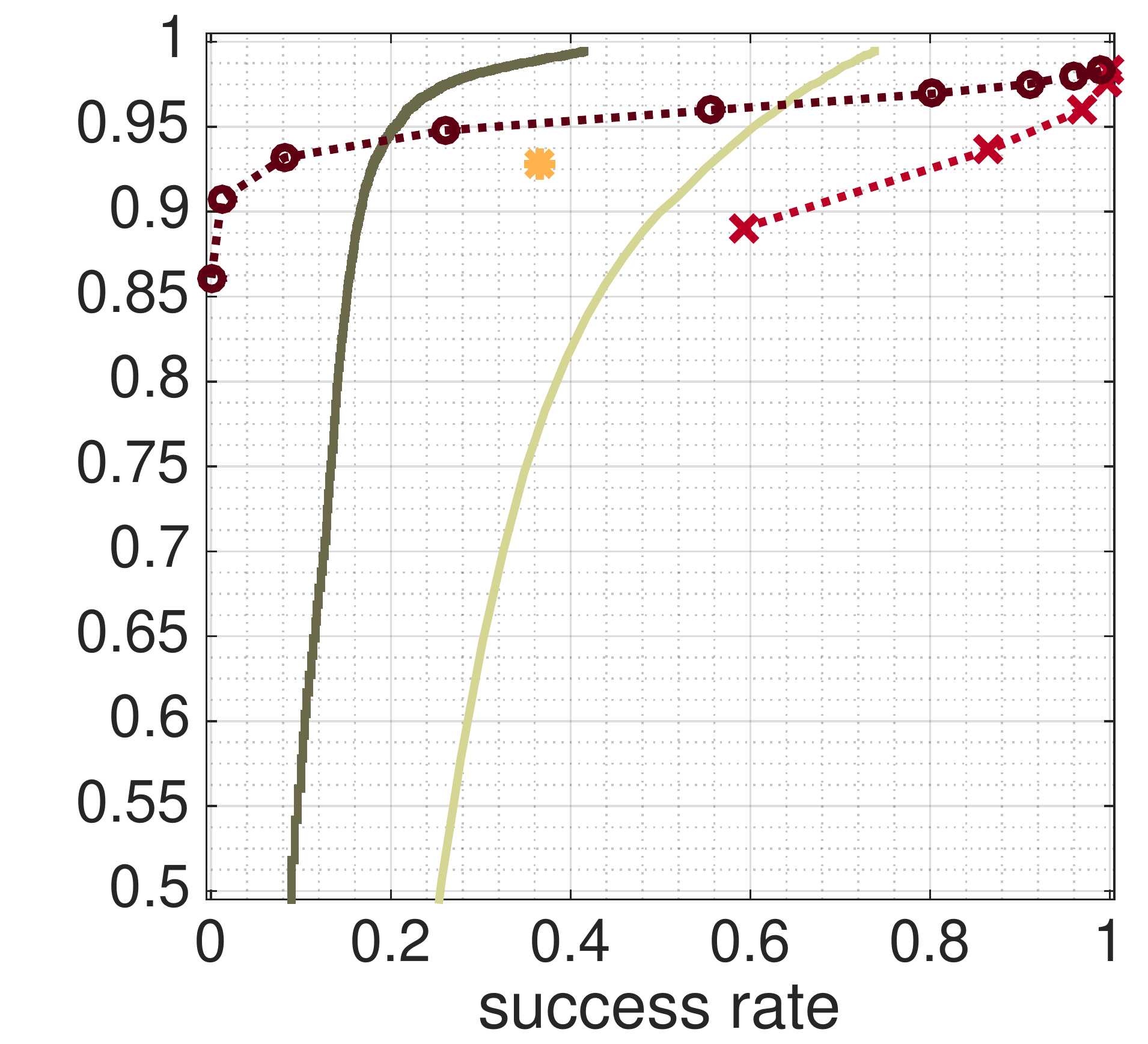}
    \caption{\gtsrb{}}
  \end{subfigure}
  \caption{\label{fig:wboxcomp}Comparison of defenses' performance in the
    \emph{white-box setting}. The \lpnorm{\infty}-norm of perturbations was set to
    $\epsilon{}=0.3$ for \mnist{} and $\epsilon{}=\frac{8}{255}$ for \cifarten{}
    and \gtsrb{}. The dotted lines connecting the \nML{} markers were added to
    help visualize trends, but do not correspond to actual operating points.}
\end{figure*}

\figref{fig:wboxcomp} shows the results. One can see that, depending on the
dataset, \nML{} outperformed other defenses, or achieved comparable performance
to the leading defenses. For \gtsrb{}, \nML{} significantly outperformed other
defenses: \nML[18] achieved a benign accuracy of 86.01\%--93.19\% at attack
success-rates $\le$8.20\%. No other defense achieved comparable benign accuracy
for such low attack success-rates. We hypothesize that \nML{} was particularly
successful for \gtsrb{}, since the dataset contains a relatively large number of
classes, and so there was a large space from which derangements for topology
manipulation could be drawn. The choice of the leading defense for \mnist{} and
\cifarten{} is less clear (some defenses achieved slightly higher benign
accuracy, while others were slightly better at preventing attacks), and depends
on the need to balance benign accuracy and resilience to attacks at
deployment time. For example, \nML[18] was slightly better at preventing
attacks against \mnist{} than \advpgd{} (1.30\% vs.\ 3.70\% attack
success-rate), but \advpgd{} achieved slightly higher accuracy for the same
\nML[18] operating point (99.25\% vs. 95.22\%).

\parheading{\lpnorm{2}-norm Attacks}
The previous experiments showed that \nML{} ensembles could resist
\lpnorm{\infty}-based attacks in various settings. We performed a preliminary
exploration using \mnist{} to assess whether \nML{} could also prevent
\lpnorm{2}-based attacks. Specifically, we trained 18 topologically manipulated
\dnn{}s to construct \nML{} ensembles. The training process was the same as
before, except that we projected adversarial perturbations to the \lpnorm{2} balls
around benign samples when performing \pgd{} to produce adversarial examples for
$\tilde{D}$. We created adversarial examples with $\epsilon\in\{0.5, 1, 2, 3\}$,
as we aimed to prevent adversarial examples with $\epsilon\le{}3$, following prior
work~\cite{Madry17AdvTraining}. The resulting topologically manipulated \dnn{}s
were accurate (an average accuracy of 98.39\%$\pm$0.55\%).

\begin{figure}
  \centering
  \includegraphics[width=0.6\columnwidth]{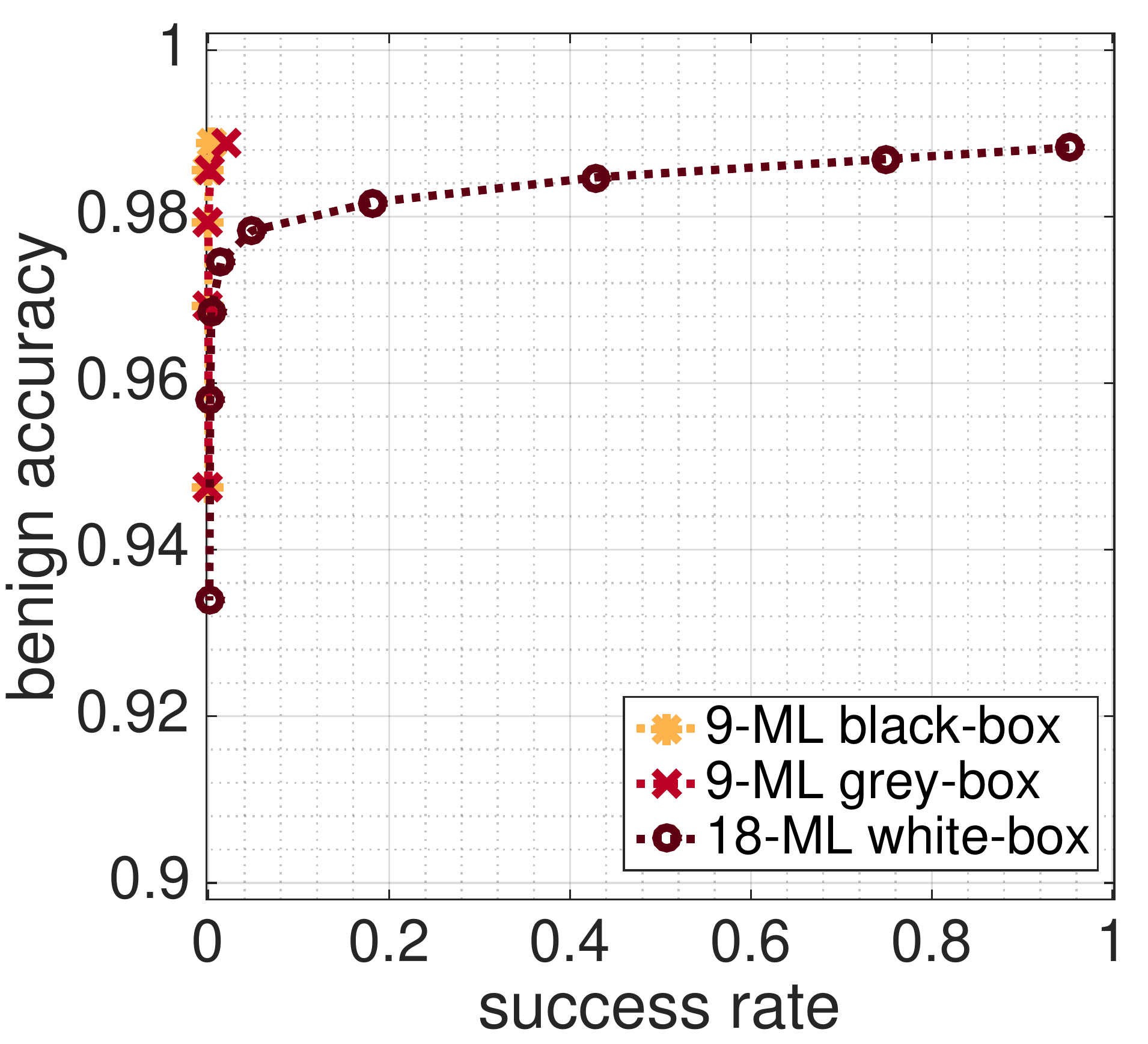}
  \caption{\label{fig:l2attacks}Performance of \mnist{} \nML{} ensembles against
    \lpnorm{2}-norm \pgd{} attacks with $\epsilon{}=3$.}
\end{figure}

sing the models that we trained, we constructed \nML{} ensembles of different sizes
and evaluated attacks in black-, grey-, and white-box settings. The
evaluation was exactly the same as before, except that we use \lpnorm{2}-based
\pgd{} with $\epsilon{}=3$ to produce adversarial
examples. \figref{fig:nml:l2attacks} summarizes the results for \nML[9] in black-
and grey-box settings, and for \nML[18] in a white-box setting. It can be seen
that \nML{} was effective at preventing \lpnorm{2}-norm attacks while
maintaining high benign accuracy. For example, \nML[9] was able to achieve 98.56\%
accuracy at $\sim$0\% success rates for black- and grey-box attacks,
and \nML[18] was able to achieve 97.46\% accuracy at a success rate of 1.40\%
for white-box attacks.

\parheading{Overhead}
Of course, as \nML{} requires us to run several \dnn{}s for inference instead of
only one, using \nML{} comes at the cost of increased inference time at
deployment. Now we show that the overhead is relatively small, especially
compared to \lid{} and \nic{}.

To measure the inference time, we sampled 1024 test samples from every dataset
and classified them in batches of 32 using the defended classifiers. We used 32
because it is common to use for inspecting the behavior of
\dnn{}s~\cite{Harlap19Pipedream}, but the trends in the time estimates remained
the same with other batch sizes. We ran the measurements on a machine equipped
with 64GB of memory and 2.50GHz Intel Xeon CPU using a single NVIDIA Tesla P100
GPU.

\begin{table}
  \centering
  \begin{tabular}{@{}r | c | p{0.75em}@{\hspace{1pt}}r @{\hspace{0.9em}} p{0.75em}@{\hspace{1pt}}r @{\hspace{0.9em}} p{0.75em}@{\hspace{1pt}}r @{\hspace{0.9em}} p{0.75em}@{\hspace{1pt}}r @{\hspace{0.9em}} p{0.75em}@{\hspace{1pt}}r@{}}
    \toprule
    \emph{Dataset} & \parbox[b]{4.5em}{\emph{Standard} / \emph{\advpgd}} &
    \multicolumn{2}{c}{\emph{\nML[5]}} &
    \multicolumn{2}{c}{\emph{\nML[9]}} &
    \multicolumn{2}{c}{\emph{\nML[18]}} &
    \multicolumn{2}{c}{\emph{\lid{}}} &
    \multicolumn{2}{c}{\emph{\nic{}}} \\ \midrule
    \mnist{}    & 0.15ms & $\times$ & 1.93 & $\times$ & 2.80 & $\times$ &  4.73 & $\times$ & 943.47 & $\times$ &  601.07 \\
    \cifarten{} & 3.72ms & $\times$ & 3.57 & $\times$ & 6.07 & $\times$ & 12.07 & $\times$ & 551.88 & $\times$ &  581.51 \\
    \gtsrb{}    & 0.68ms & $\times$ & 5.35 & $\times$ & 8.26 & $\times$ & 16.41 & $\times$ & 852.04 & $\times$ & 1457.26 \\
    \bottomrule
  \end{tabular}
  \caption{\label{tab:infertime}Defenses' overhead at inference time. The second
    column reports the average inference time in milliseconds for batches
    containing 32 samples for a single (standard or adversarially trained)
    \dnn{}. The columns to its right report the overhead of defenses with
    respect to using a single \dnn{} for inference.}
\end{table}

The results are shown in \tabref{tab:infertime}. \advpgd{} did not incur time
overhead at inference due to producing a single \dnn{} to be used for
inference. Compared to using a single \dnn{}, inference time with \nML{}
increased with $\nvariants{}$. At the same time the increase was often sublinear in
$\nvariants{}$. For example, for \nML[18], the inference time increased $\times$4.73
for \mnist{}, $\times$12.07 for \cifarten{}, and $\times$16.41 for
\gtsrb{}. Moreover, the increase was significantly less dramatic than for \lid{}
($\times$551.88--943.47 increase) and \nic{} ($\times$581.51--1,457.26
increase). There, we observed that the main bottlenecks were computing the \lid{}
statistics for \lid{}, and classification with \ocsvm{}s for \nic{}.

\section{Discussion}
\label{sec:discuss}

Our experiments demonstrated the effectiveness and efficiency of \nML{}
against \lpnorm{\infty} and \lpnorm{2} attacks in various settings for
three datasets.  Still, there are some limitations and practical
considerations to take into account when deploying \nML{}. We discuss
these below.


\parheading{Limitations}
Our experiments evaluated \nML{} against \lpnorm{2} and \lpnorm{\infty} attacks,
as is typical in the literature (e.g.,~\cite{Madry17AdvTraining,
  Shafahi19PGD}). However, in reality, attackers can use other perturbation
types to evade \nML{} (e.g., adding patterns to street signs to evade
recognition~\cite{Evtimov17Signs}). Conceptually, it should be possible to train
\nML{} to defend against such perturbation types. We defer the evaluation to
future work.

As opposed to using a single \ml{} algorithm for inference (e.g., one standard
\dnn{}), \nML{} requires using \nvariants{} \dnn{}s. As a result, more compute
resources and more time are needed to make inferences with \nML{}. This may make
it challenging to deploy \nML{} in settings where compute resources are scarce
and close to real-time feedback is expected (e.g., face-recognition on mobile
phones). Nonetheless, it is important to highlight that \nML{} is remarkably
faster at making inferences than state-of-the-art methods for detecting
adversarial examples~\cite{Ma18LID, Ma19Detection}, as our experiments showed.

Currently, perhaps the most notable weakness of \nML{} is that it is limited to
scenarios where the number of classes, $\nclasses{}$, is large. In cases where
$\nclasses{}$ is small, one cannot draw many distinct derangements to train
\dnn{}s with different topologies with which to construct \nML{} ensembles. For example,
when there are two classes ($\nclasses=2$), there is only one derangement that
one can use to train a topologically manipulated \dnn{} (remember that $!2=1$),
and so it is not possible to construct an ensemble containing $\nvariants\ge 2$
\dnn{}s with distinct topologies. A possible solution 
is to find a new method that does not require derangements to topologically
manipulate \dnn{}s. We plan to pursue this direction in future work.



\parheading{Practical considerations}
\ml{} systems often take actions based on inferences made by \ml{}
algorithms. For example, a biometric system may give or deny access to users
based on the output of a face-recognition \dnn{}; an autonomous vehicle may
change the driving direction, accelerate, stop, or slow down based on the output
of a \dnn{} for pedestrian detection; and an anti-virus program may delete or quarantine
a file based on the output of an \ml{} algorithm for malware detection. 
This raises the question of how should a system that uses \nML{} for inference
react when \nML{} flags an input as adversarial.

We have a couple of suggestions for courses of action that are applicable to
different settings. One possibility is to fall back to a more expensive, but less
error prone, classification mechanism. For example, if an \nML{} ensemble is
used for face recognition and it flags an input as adversarial, a security guard
may be called to identify the person, and possibly override the output of the
ensemble. This solution is viable when the time and resources to use an
expensive classification mechanism are available. Another possibility is to
resample the input, or classify a transformed variant of the input, to increase
the confidence in the detection or to correct the classification result. For
example, if an \nML{} ensemble that is used for face recognition on a mobile phone
detects an input as adversarial, the user may be asked to reattempt identifying
herself using a new image. In this case, because the benign accuracy of \nML{} is
high and the attack success rate is low, a benign user is likely to succeed at
identifying, while an attacker is likely to be detected.

\section{Conclusion}
\label{sec:conclude}

This paper presents \nML{}, a defense that is inspired by \nvariants-version
programming, to defend against adversarial examples. \nML{} uses ensembles of
\dnn{}s to classify inputs by a majority vote (when a large number of \dnn{}s
agree) and to detect adversarial examples (when the \dnn{}s disagree). To ensure
that the ensembles have high accuracy on benign samples while also defending
against adversarial examples, we train the \dnn{}s using a novel technique
(\emph{topology manipulation}) which allows one to specify how
adversarial examples should be classified by the \dnn{} at inference time.
Our experiments using two perturbation types (ones with bounded \lpnorm{2}-
and \lpnorm{\infty}-norms) and three datasets (\mnist{}, \cifarten{}, and
\gtsrb{}) in black-, grey-, and white-box settings showed that \nML{} is an
effective and efficient defense. In particular, \nML{} roughly retains the benign
accuracies of state-of-the-art \dnn{}s, while providing more resilience to
attacks than the best defenses known to date and making inferences faster than
most.

\ifpublishable
\section*{Acknowledgments}

This work was supported in part
by the Multidisciplinary University Research Initiative (MURI) Cyber
Deception grant; by NSF grants 1801391 and 1801494; by the National
Security Agency under Award No.\ H9823018D0008; by gifts from Google
and Nvidia, and from Lockheed Martin and NATO through Carnegie Mellon
CyLab; and by a CyLab Presidential Fellowship and a NortonLifeLock
Research Group Fellowship.
\fi

\bibliographystyle{abbrv}
\bibliography{cited}

\begin{thebibliography}{10}

\bibitem{Abadi16TensorFlow}
M.~Abadi, P.~Barham, J.~Chen, Z.~Chen, A.~Davis, J.~Dean, M.~Devin,
  S.~Ghemawat, G.~Irving, M.~Isard, M.~Kudlur, J.~Levenberg, R.~Monga,
  S.~Moore, D.~G. Murray, B.~Steiner, P.~A. Tucker, V.~Vasudevan, P.~Warden,
  M.~Wicke, Y.~Yu, and X.~Zheng.
\newblock Tensorflow: {A} system for large-scale machine learning.
\newblock In {\em Proc.\ OSDI}, 2016.

\bibitem{Abbasi17Ensemble}
M.~Abbasi and C.~Gagn{\'e}.
\newblock Robustness to adversarial examples through an ensemble of
  specialists.
\newblock In {\em Proc.\ ICLRW}, 2017.

\bibitem{Athalye18BeatCVPR}
A.~Athalye and N.~Carlini.
\newblock On the robustness of the {CVPR} 2018 white-box adversarial example
  defenses.
\newblock {\em arXiv preprint arXiv:1804.03286}, 2018.

\bibitem{Athalye18Attack}
A.~Athalye, N.~Carlini, and D.~Wagner.
\newblock Obfuscated gradients give a false sense of security: {C}ircumventing
  defenses to adversarial examples.
\newblock In {\em Proc.\ ICML}, 2018.

\bibitem{Avizienis85nVP}
A.~Avizienis.
\newblock The {N}-version approach to fault-tolerant software.
\newblock {\em IEEE Transactions on software engineering}, (12):1491--1501,
  1985.

\bibitem{Baluja17ATNs}
S.~Baluja and I.~Fischer.
\newblock Adversarial transformation networks: Learning to generate adversarial
  examples.
\newblock In {\em Proc.\ AAAI}, 2018.

\bibitem{Biggio13Evasion}
B.~Biggio, I.~Corona, D.~Maiorca, B.~Nelson, N.~{\v{S}}rndi{\'c}, P.~Laskov,
  G.~Giacinto, and F.~Roli.
\newblock Evasion attacks against machine learning at test time.
\newblock In {\em Proc.\ ECML PKDD}, 2013.

\bibitem{Brendel18DBA}
W.~Brendel, J.~Rauber, and M.~Bethge.
\newblock Decision-based adversarial attacks: {R}eliable attacks against
  black-box machine learning models.
\newblock In {\em Proc.\ ICLR}, 2018.

\bibitem{Brunner18BBox}
T.~Brunner, F.~Diehl, M.~T. Le, and A.~Knoll.
\newblock Guessing smart: Biased sampling for efficient black-box adversarial
  attacks.
\newblock In {\em Proc.\ ICCV}, 2019.

\bibitem{Carlini17Bypass}
N.~Carlini and D.~Wagner.
\newblock Adversarial examples are not easily detected: Bypassing ten detection
  methods.
\newblock In {\em Proc.\ AISec}, 2017.

\bibitem{Carlini17Robustness}
N.~Carlini and D.~Wagner.
\newblock Towards evaluating the robustness of neural networks.
\newblock In {\em Proc.\ IEEE S{\&}P}, 2017.

\bibitem{Chen95nVP}
L.~Chen and A.~Avizienis.
\newblock {N}-version programming: {A} fault-tolerance approach to reliability
  of software operation.
\newblock In {\em Proc.\ ISFTC}, 1995.

\bibitem{Chen17Zoo}
P.-Y. Chen, H.~Zhang, Y.~Sharma, J.~Yi, and C.-J. Hsieh.
\newblock Zoo: {Z}eroth order optimization based black-box attacks to deep
  neural networks without training substitute models.
\newblock In {\em Proc.\ AISec}, 2017.

\bibitem{Cohen19RandSmooth}
J.~M. Cohen, E.~Rosenfeld, and J.~Z. Kolter.
\newblock Certified adversarial robustness via randomized smoothing.
\newblock In {\em Proc.\ ICML}, 2019.

\bibitem{Cox06nVS}
B.~Cox, D.~Evans, A.~Filipi, J.~Rowanhill, W.~Hu, J.~Davidson, J.~Knight,
  A.~Nguyen-Tuong, and J.~Hiser.
\newblock N-variant systems: A secretless framework for security through
  diversity.
\newblock In {\em Proc.\ USENIX Security}, 2006.

\bibitem{Demontis19Transfer}
A.~Demontis, M.~Melis, M.~Pintor, M.~Jagielski, B.~Biggio, A.~Oprea,
  C.~Nita-Rotaru, and F.~Roli.
\newblock Why do adversarial attacks transfer? {E}xplaining transferability of
  evasion and poisoning attacks.
\newblock In {\em Proc.\ USENIX Security}, 2019.

\bibitem{Dong19TI}
Y.~Dong, T.~Pang, H.~Su, and J.~Zhu.
\newblock Evading defenses to transferable adversarial examples by
  translation-invariant attacks.
\newblock In {\em Proc.\ CVPR}, 2019.

\bibitem{Evtimov17Signs}
I.~Evtimov, K.~Eykholt, E.~Fernandes, T.~Kohno, B.~Li, A.~Prakash, A.~Rahmati,
  and D.~Song.
\newblock Robust physical-world attacks on machine learning models.
\newblock In {\em Proc.\ CVPR}, 2018.

\bibitem{Feinman17Detector}
R.~Feinman, R.~R. Curtin, S.~Shintre, and A.~B. Gardner.
\newblock Detecting adversarial samples from artifacts.
\newblock {\em arXiv preprint arXiv:1703.00410}, 2017.

\bibitem{Gdfllw14ExpAdv}
I.~J. Goodfellow, J.~Shlens, and C.~Szegedy.
\newblock Explaining and harnessing adversarial examples.
\newblock In {\em Proc.\ ICLR}, 2015.

\bibitem{Grosse17Detector}
K.~Grosse, P.~Manoharan, N.~Papernot, M.~Backes, and P.~McDaniel.
\newblock On the (statistical) detection of adversarial examples.
\newblock {\em arXiv preprint arXiv:1702.06280}, 2017.

\bibitem{Guo18ReformDef}
C.~Guo, M.~Rana, M.~Cisse, and L.~van~der Maaten.
\newblock Countering adversarial images using input transformations.
\newblock In {\em Proc.\ ICLR}, 2018.

\bibitem{Harlap19Pipedream}
A.~Harlap, D.~Narayanan, A.~Phanishayee, V.~Seshadri, N.~Devanur, G.~Ganger,
  and P.~Gibbons.
\newblock Pipedream: Fast and efficient pipeline parallel dnn training.
\newblock In {\em Proc.\ SOSP}, 2019.
\newblock To appear.

\bibitem{He17FoolEnsembles}
W.~He, J.~Wei, X.~Chen, N.~Carlini, and D.~Song.
\newblock Adversarial example defense: {E}nsembles of weak defenses are not
  strong.
\newblock In {\em Proc.\ WOOT}, 2017.

\bibitem{Huang19Random}
B.~Huang, Y.~Wang, and W.~Wang.
\newblock Model-agnostic adversarial detection by random perturbations.
\newblock In {\em Proc.\ IJCAI}, 2019.

\bibitem{Ilyas19Explain}
A.~Ilyas, L.~Engstrom, and A.~Madry.
\newblock Prior convictions: {B}lack-box adversarial attacks with bandits and
  priors.
\newblock In {\em Proc.\ ICLR}, 2019.

\bibitem{Kannan18ALP}
H.~Kannan, A.~Kurakin, and I.~Goodfellow.
\newblock Adversarial logit pairing.
\newblock {\em arXiv preprint arXiv:1803.06373}, 2018.

\bibitem{Kantchelian16ICML}
A.~Kantchelian, J.~Tygar, and A.~D. Joseph.
\newblock Evasion and hardening of tree ensemble classifiers.
\newblock In {\em Proc.\ ICML}, 2016.

\bibitem{Kari19EnsembleDiv}
S.~Kariyappa and M.~K. Qureshi.
\newblock Improving adversarial robustness of ensembles with diversity
  training.
\newblock {\em arXiv preprint arXiv:1901.09981}, 2019.

\bibitem{Keras15}
{Keras team}.
\newblock Keras: {T}he {P}ython deep learning library.
\newblock \url{https://keras.io/}, 2015.
\newblock Accessed on 09-30-2019.

\bibitem{KerasMNISTCNN}
{Keras team}.
\newblock {MNIST CNN}.
\newblock
  \url{https://github.com/keras-team/keras/blob/master/examples/mnist_cnn.py},
  2018.
\newblock Accessed on 09-28-2019.

\bibitem{KerasMNISTMLP}
{Keras team}.
\newblock {MNIST MLP}.
\newblock \url{https://keras.io/examples/mnist_mlp/}, 2018.
\newblock Accessed on 09-28-2019.

\bibitem{Kingma14Adam}
D.~Kingma and J.~Ba.
\newblock Adam: {A} method for stochastic optimization.
\newblock In {\em Proc.\ ICLR}, 2015.

\bibitem{Kolter17Defense}
J.~Z. Kolter and E.~Wong.
\newblock Provable defenses against adversarial examples via the convex outer
  adversarial polytope.
\newblock In {\em Proc.\ ICML}, 2018.

\bibitem{Krizhevsky09CIFAR}
A.~Krizhevsky, G.~Hinton, et~al.
\newblock Learning multiple layers of features from tiny images.
\newblock Technical report, University of Toronto, 2009.

\bibitem{Kurakin16AdvTrain}
A.~Kurakin, I.~Goodfellow, and S.~Bengio.
\newblock Adversarial machine learning at scale.
\newblock In {\em Proc.\ ICLR}, 2017.

\bibitem{MNIST}
Y.~LeCun, C.~Cortes, and C.~J. Burges.
\newblock The {MNIST} database of handwritten digits.
\newblock \url{http://yann.lecun.com/exdb/mnist/}, 1998.
\newblock Accessed on 10-01-2019.

\bibitem{Lecuyer19PixelDP}
M.~Lecuyer, V.~Atlidakis, R.~Geambasu, D.~Hsu, and S.~Jana.
\newblock Certified robustness to adversarial examples with differential
  privacy.
\newblock In {\em Proc.\ IEEE S\&P}, 2019.

\bibitem{Li18GTSRB}
J.~Li and Z.~Wang.
\newblock Real-time traffic sign recognition based on efficient {CNN}s in the
  wild.
\newblock {\em IEEE Transactions on Intelligent Transportation Systems},
  20(3):975--984, 2018.

\bibitem{Liao18ReformDef}
F.~Liao, M.~Liang, Y.~Dong, T.~Pang, J.~Zhu, and X.~Hu.
\newblock Defense against adversarial attacks using high-level representation
  guided denoiser.
\newblock In {\em Proc.\ CVPR}, 2018.

\bibitem{Liu18RSE}
X.~Liu, M.~Cheng, H.~Zhang, and C.-J. Hsieh.
\newblock Towards robust neural networks via random self-ensemble.
\newblock In {\em Proc.\ ECCV}, 2018.

\bibitem{Liu17Transfer}
Y.~Liu, X.~Chen, C.~Liu, and D.~Song.
\newblock Delving into transferable adversarial examples and black-box attacks.
\newblock In {\em Proc.\ ICLR}, 2017.

\bibitem{Lu18FoolLID}
P.-H. Lu, P.-Y. Chen, and C.-M. Yu.
\newblock On the limitation of local intrinsic dimensionality for
  characterizing the subspaces of adversarial examples.
\newblock {\em arXiv preprint arXiv:1803.09638}, 2018.

\bibitem{Ma19Detection}
S.~Ma, Y.~Liu, G.~Tao, W.-C. Lee, and X.~Zhang.
\newblock {NIC}: Detecting adversarial samples with neural network invariant
  checking.
\newblock In {\em Proc.\ NDSS}, 2019.

\bibitem{Ma18LID}
X.~Ma, B.~Li, Y.~Wang, S.~M. Erfani, S.~Wijewickrema, G.~Schoenebeck, D.~Song,
  M.~E. Houle, and J.~Bailey.
\newblock Characterizing adversarial subspaces using local intrinsic
  dimensionality.
\newblock In {\em Proc.\ ICLR}, 2018.

\bibitem{Machida19nML}
F.~Machida.
\newblock {N}-version machine learning models for safety critical systems.
\newblock In {\em Proc.\ DSN DSMLW}, 2019.

\bibitem{Madry17AdvTraining}
A.~Madry, A.~Makelov, L.~Schmidt, D.~Tsipras, and A.~Vladu.
\newblock Towards deep learning models resistant to adversarial attacks.
\newblock In {\em Proc.\ ICLR}, 2018.

\bibitem{Meng17Magnet}
D.~Meng and H.~Chen.
\newblock Magnet: {A} two-pronged defense against adversarial examples.
\newblock In {\em Proc.\ CCS}, 2017.

\bibitem{Metzen17Detector}
J.~H. Metzen, T.~Genewein, V.~Fischer, and B.~Bischoff.
\newblock On detecting adversarial perturbations.
\newblock In {\em Proc.\ ICLR}, 2017.

\bibitem{Mirman18Defense}
M.~Mirman, T.~Gehr, and M.~Vechev.
\newblock Differentiable abstract interpretation for provably robust neural
  networks.
\newblock In {\em Proc.\ ICML}, 2018.

\bibitem{Moosavi16DeepFool}
S.-M. Moosavi-Dezfooli, A.~Fawzi, and P.~Frossard.
\newblock Deep{F}ool: {A} simple and accurate method to fool deep neural
  networks.
\newblock In {\em Proc.\ CVPR}, 2016.

\bibitem{Pang18DetectRCE}
T.~Pang, C.~Du, Y.~Dong, and J.~Zhu.
\newblock Towards robust detection of adversarial examples.
\newblock In {\em Proc.\ NeurIPS}, 2018.

\bibitem{Pang19Ensembles}
T.~Pang, K.~Xu, C.~Du, N.~Chen, and J.~Zhu.
\newblock Improving adversarial robustness via promoting ensemble diversity.
\newblock In {\em Proc.\ ICML}, 2019.

\bibitem{Papernot18DkNN}
N.~Papernot and P.~McDaniel.
\newblock Deep k-nearest neighbors: {T}owards confident, interpretable and
  robust deep learning.
\newblock {\em arXiv preprint arXiv:1803.04765}, 2018.

\bibitem{Papernot16Transferability}
N.~Papernot, P.~McDaniel, and I.~Goodfellow.
\newblock Transferability in machine learning: {F}rom phenomena to black-box
  attacks using adversarial samples.
\newblock {\em arXiv preprint arXiv:1605.07277}, 2016.

\bibitem{Papernot17Blackbox}
N.~Papernot, P.~McDaniel, I.~Goodfellow, S.~Jha, Z.~B. Celik, and A.~Swami.
\newblock Practical black-box attacks against machine learning.
\newblock In {\em Proc.\ AsiaCCS}, 2017.

\bibitem{Papernot16Limitations}
N.~Papernot, P.~McDaniel, S.~Jha, M.~Fredrikson, Z.~B. Celik, and A.~Swami.
\newblock The limitations of deep learning in adversarial settings.
\newblock In {\em Proc. IEEE Euro S{\&}P}, 2016.

\bibitem{Qin19Speech}
Y.~Qin, N.~Carlini, I.~Goodfellow, G.~Cottrell, and C.~Raffel.
\newblock Imperceptible, robust, and targeted adversarial examples for
  automatic speech recognition.
\newblock In {\em Proc.\ ICML}, 2019.

\bibitem{Raghu18Certified}
A.~Raghunathan, J.~Steinhardt, and P.~S. Liang.
\newblock Semidefinite relaxations for certifying robustness to adversarial
  examples.
\newblock In {\em Proc.\ NeurIPS}, 2018.

\bibitem{Salman19Robust}
H.~Salman, G.~Yang, J.~Li, P.~Zhang, H.~Zhang, I.~Razenshteyn, and S.~Bubeck.
\newblock Provably robust deep learning via adversarially trained smoothed
  classifiers.
\newblock In {\em Proc.\ NeurIPS}, 2019.
\newblock To appear.

\bibitem{Salman19Convex}
H.~Salman, G.~Yang, H.~Zhang, C.-J. Hsieh, and P.~Zhang.
\newblock A convex relaxation barrier to tight robustness verification of
  neural networks.
\newblock In {\em Proc.\ NeurIPS}, 2019.
\newblock To appear.

\bibitem{Samangouei18DefGAN}
P.~Samangouei, M.~Kabkab, and R.~Chellappa.
\newblock Defense-{GAN}: {P}rotecting classifiers against adversarial attacks
  using generative models.
\newblock In {\em Proc.\ ICLR}, 2018.

\bibitem{Schon18AdvSound}
L.~Sch{\"o}nherr, K.~Kohls, S.~Zeiler, T.~Holz, and D.~Kolossa.
\newblock Adversarial attacks against automatic speech recognition systems via
  psychoacoustic hiding.
\newblock In {\em Proc.\ NDSS}, 2019.

\bibitem{Sen19Lp}
A.~Sen, X.~Zhu, L.~Marshall, and R.~Nowak.
\newblock Should adversarial attacks use pixel p-norm?
\newblock {\em arXiv preprint arXiv:1906.02439}, 2019.

\bibitem{Sengupta19MTDeep}
S.~Sengupta, T.~Chakraborti, and S.~Kambhampati.
\newblock {MTDeep}: {M}oving target defense to boost the security of deep
  neural nets against adversarial attacks.
\newblock In {\em Proc.\ GameSec}, 2019.

\bibitem{Sermanet11GTSRB}
P.~Sermanet and Y.~LeCun.
\newblock Traffic sign recognition with multi-scale convolutional networks.
\newblock In {\em Proc.\ IJCNN}, 2011.

\bibitem{Shafahi19PGD}
A.~Shafahi, M.~Najibi, A.~Ghiasi, Z.~Xu, J.~Dickerson, C.~Studer, L.~S. Davis,
  G.~Taylor, and T.~Goldstein.
\newblock Adversarial training for free!
\newblock In {\em Proc.\ NeurIPS}, 2019.
\newblock To appear.

\bibitem{Sharif18Lp}
M.~Sharif, L.~Bauer, and M.~K. Reiter.
\newblock On the suitability of lp-norms for creating and preventing
  adversarial examples.
\newblock In {\em Proc.\ CVPRW}, 2018.

\bibitem{Sharif16AdvML}
M.~Sharif, S.~Bhagavatula, L.~Bauer, and M.~K. Reiter.
\newblock Accessorize to a crime: {R}eal and stealthy attacks on
  state-of-the-art face recognition.
\newblock In {\em Proc.\ CCS}, 2016.

\bibitem{Spring15AllConv}
J.~T. Springenberg, A.~Dosovitskiy, T.~Brox, and M.~Riedmiller.
\newblock Striving for simplicity: The all convolutional net.
\newblock In {\em Proc.\ ICLR}, 2015.

\bibitem{Srini18ReformDef}
V.~Srinivasan, A.~Marban, K.-R. M{\"u}ller, W.~Samek, and S.~Nakajima.
\newblock Counterstrike: Defending deep learning architectures against
  adversarial samples by {L}angevin dynamics with supervised denoising
  autoencoder.
\newblock {\em arXiv preprint arXiv:1805.12017}, 2018.

\bibitem{Stallkamp12GTSRB}
J.~Stallkamp, M.~Schlipsing, J.~Salmen, and C.~Igel.
\newblock Man vs. computer: Benchmarking machine learning algorithms for
  traffic sign recognition.
\newblock {\em Neural Networks}, 32:323--332, 2012.

\bibitem{Strauss17Ensemble}
T.~Strauss, M.~Hanselmann, A.~Junginger, and H.~Ulmer.
\newblock Ensemble methods as a defense to adversarial perturbations against
  deep neural networks.
\newblock {\em arXiv preprint arXiv:1709.03423}, 2017.

\bibitem{Szegedy13NNsProps}
C.~Szegedy, W.~Zaremba, I.~Sutskever, J.~Bruna, D.~Erhan, I.~J. Goodfellow, and
  R.~Fergus.
\newblock Intriguing properties of neural networks.
\newblock In {\em Proc.\ ICLR}, 2014.

\bibitem{Tian17GTSRB}
L.~Tian.
\newblock Traffic sign recognition using {CNN} with learned color and spatial
  transformation.
\newblock
  \url{https://github.com/hello2all/GTSRB_Keras_STN/blob/master/conv_model.py},
  2017.
\newblock Accessed on 09-28-2019.

\bibitem{Tramer18Ensemble}
F.~Tram{\`e}r, A.~Kurakin, N.~Papernot, I.~Goodfellow, D.~Boneh, and
  P.~McDaniel.
\newblock Ensemble adversarial training: Attacks and defenses.
\newblock In {\em Proc.\ ICLR}, 2018.

\bibitem{Vijay19Cascades}
D.~Vijaykeerthy, A.~Suri, S.~Mehta, and P.~Kumaraguru.
\newblock Hardening deep neural networks via adversarial model cascades.
\newblock 2019.

\bibitem{Wang19Mutation}
J.~Wang, G.~Dong, J.~Sun, X.~Wang, and P.~Zhang.
\newblock Adversarial sample detection for deep neural network through model
  mutation testing.
\newblock In {\em Proc.\ ICSE}, 2019.

\bibitem{Wang19HRS}
X.~Wang, S.~Wang, P.-Y. Chen, Y.~Wang, B.~Kulis, X.~Lin, and P.~Chin.
\newblock Protecting neural networks with hierarchical random switching:
  {T}owards better robustness-accuracy trade-off for stochastic defenses.
\newblock {\em arXiv preprint arXiv:1908.07116}, 2019.

\bibitem{Xiao18GANAttack}
C.~Xiao, B.~Li, J.-Y. Zhu, W.~He, M.~Liu, and D.~Song.
\newblock Generating adversarial examples with adversarial networks.
\newblock In {\em Proc.\ IJCAI}, 2018.

\bibitem{Xie18Denoise}
C.~Xie, Y.~Wu, L.~van~der Maaten, A.~Yuille, and K.~He.
\newblock Feature denoising for improving adversarial robustness.
\newblock {\em arXiv preprint arXiv:1812.03411}, 2018.

\bibitem{Xie19ID}
C.~Xie, Z.~Zhang, Y.~Zhou, S.~Bai, J.~Wang, Z.~Ren, and A.~L. Yuille.
\newblock Improving transferability of adversarial examples with input
  diversity.
\newblock In {\em Proc.\ CVPR}, 2019.

\bibitem{Xu19nvDNN}
H.~Xu, Z.~Chen, W.~Wu, Z.~Jin, S.-y. Kuo, and M.~Lyu.
\newblock {NV-DNN}: {T}owards fault-tolerant {DNN} systems with {N}-version
  programming.
\newblock In {\em Proc.\ DSN DSMLW}, 2019.

\bibitem{Xu18Squeeze}
W.~Xu, D.~Evans, and Y.~Qi.
\newblock Feature squeezing: Detecting adversarial examples in deep neural
  networks.
\newblock In {\em Proc.\ NDSS}, 2018.

\bibitem{Zagoruyko16WRN}
S.~Zagoruyko and N.~Komodakis.
\newblock Wide residual networks.
\newblock In {\em Proc.\ BMVC}, 2016.

\bibitem{Zeiler12Adadelta}
M.~D. Zeiler.
\newblock Adadelta: {A}n adaptive learning rate method.
\newblock {\em arXiv preprint arXiv:1212.5701}, 2012.

\bibitem{Zeng19MVP}
Q.~Zeng, J.~Su, C.~Fu, G.~Kayas, L.~Luo, X.~Du, C.~C. Tan, and J.~Wu.
\newblock A multiversion programming inspired approach to detecting audio
  adversarial examples.
\newblock In {\em Proc.\ DSN}, 2019.

\end{thebibliography}

\end{document}